\documentclass[runningheads]{llncs}
\usepackage[T1]{fontenc}
\usepackage[numbers,sort&compress]{natbib}
\usepackage{booktabs}
\usepackage{subfigure}
\usepackage{array}
\usepackage{float}
\usepackage{placeins}
\usepackage{tikz}
\usetikzlibrary{arrows.meta,calc,fit}
\usepackage{microtype}
\usepackage{amsmath}
\usepackage{amssymb}
\usepackage[linesnumbered,ruled,vlined]{algorithm2e}
\usepackage{multirow}
\usepackage{graphicx}
\usepackage{algpseudocode}
\usepackage{nomencl}

\begin{document}
\raggedbottom
\emergencystretch=1.5em
\title{GFGE: Unifying Explainable AI Methods through an Interpretation Framework}
\titlerunning{GFGE: A Framework for XAI Methods}
%

\author{Jinfeng Zhong\inst{1}}
\authorrunning{Zhong}
%
\institute{LIASD, Université Paris 8, 93200 Saint-Denis, France\\
\email{jinfeng.zhong@univ-paris8.fr}}

\maketitle              
\begin{abstract}
Explainable artificial intelligence (XAI) encompasses methods that draw on
different sources of information and address different explanatory needs.
A common framework is needed to describe how this information becomes
evidence and is communicated as an explanation for a particular recipient.
We propose the General Framework for Generating Explanations (GFGE), grounded
in interpretative frameworks and the complementary activities of
\emph{sense-reading} and \emph{sense-giving}. Its conceptual foundation is the
Interpret/Explain Schema (IES), which connects an analyst's interpretation of
system evidence, the communication of a selected account, and the recipient's
interpretation of that account. GFGE operationalises this schema through five
roles: data interpretation, model interpretation, output interpretation,
optional post-hoc analysis, and aggregation. A role-typed operation graph
records method-specific dependencies, while evidence records retain the
sources, assumptions, and limitations of explanatory claims. The explanatory
question, audience, and context guide the procedure. We instantiate GFGE for
attribution, surrogate, counterfactual, concept and prototype, intrinsic rule,
argumentation, and language-model methods. These instantiations show how
intrinsic, post-hoc, and hybrid workflows can be represented through the same
roles while preserving their distinct evidential requirements. GFGE provides
a common basis for analysing explanation workflows, tracing communicated
claims to their evidence, and identifying unresolved explanatory dependencies.

\keywords{Explainable artificial intelligence \and Interpretation framework \and Explanation generation \and Counterfactual explanations}
\end{abstract}

\section{Introduction}

AI systems increasingly influence decisions that directly affect individuals, yet
users often struggle to understand why these systems behave as they do. Consider
a loan applicant who receives a rejection from an automated credit-scoring
system, or a patient who is told that a clinical model predicts a high risk of
disease. In such situations, people do not only want to know \emph{what} the
system decided but also \emph{why} the decision was made, whether it is fair,
and what could be changed to obtain a different outcome. These examples reveal a
practical tension: although modern predictive models achieve high accuracy,
their internal reasoning is not immediately interpretable, and existing
explanation techniques draw on different evidence and answer different
questions. A shared account must preserve these differences while showing how
each technique contributes to an explanation for its intended audience.

As AI-based decision support becomes increasingly embedded across domains such
as healthcare, finance, transportation, and online services, questions of
transparency, accountability, and trustworthiness have become central to their
design and deployment \citep{Samek2021Review,Vilone2021Notions,molnar2020interpretable}. Users,
developers, and regulators now expect AI systems not only to deliver accurate
predictions but also to justify their outputs in meaningful and understandable
ways. This demand is reinforced by legal and ethical frameworks, including the
GDPR and the EU AI Act. Their transparency and oversight requirements depend on
the application context; they do not require the same explanatory artefact for
every AI output \citep{Wachter2017,EUAIAct2024}.

Yet despite broad agreement on the \emph{importance} of explanations, the
concepts of \emph{interpretability} and \emph{explainability} remain remarkably
fragmented. Existing literature provides divergent definitions: some works
attempt to distinguish interpretability from explainability
\citep{lipton2018mythos,doshi2017towards,montavon2018methods}, while others use the terms interchangeably
\citep{miller2019explanation,molnar2020interpretable,arrieta2020explainable}. These usages are often rooted in
discipline-specific intuition rather than shared principles, leading to
terminological ambiguity across computer science, philosophy, and cognitive
science. This lack of conceptual clarity complicates the formulation of rigorous
research questions, obscures the relationships among explanatory methods, and
limits our ability to evaluate and compare explainable AI (XAI) systems in a
coherent manner.

\bigskip
Our primary aim is to unify the generation of explanations across heterogeneous
XAI methods. To that end, this paper makes four contributions.

\textbf{(1) A conceptual distinction between interpretation and explanation.}
Drawing on linguistic analysis, philosophy of explanation, cognitive science,
and knowledge management, we disentangle the notions of interpretation
(sense-reading) and explanation (sense-giving). We give working definitions
that distinguish a person's act of making sense from a computational operation
that extracts candidate evidence. Explanatory questions include why, why-not,
what-if, how-to, and global-pattern questions.

\textbf{(2) The Interpret/Explain Schema (IES) for AI systems.}
Building on these definitions, we propose a schema that characterises how
interpretation and explanation emerge within the data--model--output pipeline of
an AI system. The IES clarifies the flow of information between developers,
systems, and end-users, operationalising abstract conceptual distinctions into a
computational structure.

\textbf{(3) The General Framework for Generating Explanations (GFGE).}
We introduce GFGE, a modular and model-agnostic framework that instantiates the
IES as a role-typed operation graph for describing explanation generation. GFGE is
parameterised by components for data interpretation, model interpretation, output
interpretation, post-hoc interpretation, and aggregation. Unlike many existing
explanation methods, GFGE is not tied to any specific model family (e.g., linear
models, trees, neural networks) or any specific post-hoc technique (e.g., LIME or
SHAP). Instead, these components function as interchangeable modules, making
GFGE compatible with both intrinsic and post-hoc explanation paradigms.

\textbf{(4) Cross-family instantiation and analysis.}
To examine the descriptive scope of GFGE, we instantiate it with a
diverse range of explanation paradigms: post-hoc attribution techniques (e.g.,
LIME, SHAP), methods built upon attribution models (e.g., SHAP-guided
counterfactuals, recourse methods), surrogate-based explanations, prototype and
concept-based approaches, and intrinsically interpretable techniques such as
argumentation-based models and language-model explanations
\citep{ribeiro2016should,lundberg2017unified,albini2022counterfactual,
chen2019looks,nauta2021neural,zhong2022a3r,zhong2025argumentation,
chuang2026faithlm}. A comparative table specifies the GFGE roles used by
each family, including hybrid procedures that combine several operations.
We then test the mapping rules on three boundary cases and compare what two
nearby frameworks and GFGE record for the same methods.

\bigskip
This is a conceptual framework paper supported by a structured, purposive
review and author-coded method analyses. The comparison tests descriptive
expressiveness, not the empirical quality of generated explanations.

\section{Literature Review}\label{LR}

This section reviews existing work on explanation in AI with the specific goal
of identifying conceptual and structural gaps that motivate the definitions
and framework proposed in this paper.

\paragraph{Review Scope and Selection.}
This is a structured narrative review, not a systematic or exhaustive survey.
We selected work that (i) defines interpretation or explanation, (ii) offers
cross-method taxonomy, stakeholder, or selection frameworks, or (iii) provides
a method whose evidence flow creates a boundary case for GFGE. For the
framework comparison, we use the broad method-trait taxonomy of
Schwalbe and Finzel~\citep{schwalbe2024taxonomy} and the stakeholder/lifecycle roadmap of
Wang et al.~\citep{wang2024roadmap}; the other reviews in
Table~\ref{tab:survey_comparison} provide context. For the boundary analysis,
we use an external concept model, a perturbation reliability method, and an
LLM explanation method, covering model access, evidence quality, and
communication, respectively. These cases were chosen for conceptual
diversity, not by random sampling. The literature set cannot support
prevalence claims or comprehensive coverage of XAI.

\subsection{A General View}\label{sec:2.1}

As AI systems are increasingly deployed across a wide range of applications,
their decisions are expected to be not only accurate but also intelligible to
various stakeholders \citep{Samek2021Review, Vilone2021Notions, molnar2020interpretable}.
This expectation has led to a growing body of work on explainable artificial
intelligence (XAI), addressing transparency, accountability, and trust in
algorithmic decision-making.

Despite this growing attention, the terminology surrounding
\emph{interpretability} and \emph{explainability} remains inconsistent.
Some authors explicitly distinguish between the two notions
\citep{lipton2018mythos, doshi2017towards, montavon2018methods}, while others
use them interchangeably or treat the distinction as secondary
\citep{miller2019explanation, molnar2020interpretable, du2019techniques,
arrieta2020explainable, Bhatt2020}.
These differences are often grounded in discipline-specific intuitions rather
than shared theoretical principles \citep{miller2019explanation}.
As a result, key terms are used with varying meanings across the literature,
which complicates the formulation, comparison, and evaluation of explanation
methods.

This lack of conceptual clarity motivates a closer examination of how
interpretability and explainability are defined and related, and how they are
operationalised within AI systems.

\subsection{To Explain or to Interpret?}\label{sec2.2}

The relationship between interpretability and explainability has been discussed
extensively, yet no consensus has emerged.
Three broad positions can be identified in the literature.

A first line of work treats interpretability as a prerequisite for explainability,
arguing that models whose internal mechanisms are transparent can explain their
outputs by default \citep{lipton2018mythos, montavon2018methods,
guidotti2018survey, mittelstadt2019explaining, rudin2019stop}.
In this view, interpretation primarily concerns understanding how a model
operates, while explanation concerns why a particular result was produced.

A second position reverses this implication, suggesting that explainable models
are interpretable by default, but not all interpretable models yield valid
explanations \citep{gilpin2018explaining}.
Here, the emphasis is placed on the justification and correctness of explanations,
often with a causal interpretation.

A third line of work uses the terms interpretability and explainability
interchangeably, particularly in survey and applied contexts
\citep{miller2019explanation, molnar2020interpretable, arrieta2020explainable}.
In these works, interpretability may refer either to model transparency or to
the ability to explain individual predictions.

Taken together, these positions show that existing distinctions are neither uniform
nor systematically grounded, which limits their usefulness for structuring
explanation methods across models and application contexts.

\subsection{Explanation: What?}\label{sec2.3}

\begin{figure}[t]
\centering
\begingroup
\input{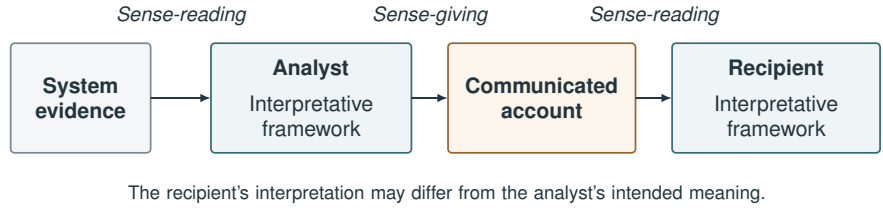}
\begin{tikzpicture}[gf base]
  \node[gf box,fill=gfSystemFill,text width=1.5cm,
    minimum width=1.85cm,minimum height=1.45cm] (evidence) at (0.95,0)
    {\textbf{System}\\\textbf{evidence}};
  \node[gf human,text width=2.3cm,
    minimum width=2.65cm,minimum height=1.45cm] (analyst) at (4,0)
    {\textbf{Analyst}\\[4pt]Interpretative\\framework};
  \node[gf account,text width=2.12cm,
    minimum width=2.48cm,minimum height=1.45cm] (account) at (7.05,0)
    {\textbf{Communicated}\\\textbf{account}};
  \node[gf human,text width=2.4cm,
    minimum width=2.75cm,minimum height=1.45cm] (recipient) at (10.15,0)
    {\textbf{Recipient}\\[4pt]Interpretative\\framework};

  \draw[gf flow] (evidence.east) -- (analyst.west);
  \draw[gf flow] (analyst.east) -- (account.west);
  \draw[gf flow] (account.east) -- (recipient.west);

  \node[font=\fontfamily{phv}\selectfont\fontsize{7.5}{9}\selectfont\itshape]
    at ($(evidence.east)!0.5!(analyst.west)+(0,1.09)$) {Sense-reading};
  \node[font=\fontfamily{phv}\selectfont\fontsize{7.5}{9}\selectfont\itshape]
    at ($(analyst.east)!0.5!(account.west)+(0,1.09)$) {Sense-giving};
  \node[font=\fontfamily{phv}\selectfont\fontsize{7.5}{9}\selectfont\itshape]
    at ($(account.east)!0.5!(recipient.west)+(0,1.09)$) {Sense-reading};

  \node[gf small,align=center,text width=11.4cm] at (5.775,-1.27)
    {The recipient's interpretation may differ from the analyst's intended meaning.};
\end{tikzpicture}
\endgroup
\caption{An analyst reads evidence through an interpretative framework, gives
an account to a recipient, and the recipient reads that account through a
different framework. This is our conceptual illustration of sense-reading and
sense-giving \citep{despres1999knowledge,arduin2015information}.}
\label{fig:reading_giving}
\end{figure}

The notion of explanation has been studied long before the advent of AI, notably
in philosophy, cognitive science, and social psychology.
Philosophical accounts emphasise causal structure
\citep{lewis1986causal, josephson1996abductive},
while cognitive and social perspectives stress the communicative and
interactional nature of explanations
\citep{malle2006mind, lombrozo2006structure, miller2019explanation}.

Across these traditions, explanations are commonly understood as having a dual
nature: they are both a \emph{process}---involving the construction and selection
of relevant information---and a \emph{product} that addresses an explanatory
question. The question may request a reason, a contrast, a hypothetical
outcome, a feasible change, or a global pattern.
Importantly, explanations are not purely objective artefacts, but are shaped by
the beliefs, expectations, and background knowledge of their recipients.

These insights align with concepts from knowledge management, where explanation
emerges from the interaction between \emph{sense-reading} (interpreting
information through an individual mental model) and \emph{sense-giving}
(assigning meaning for others) \citep{despres1999knowledge, arduin2015information}.
As illustrated in Figure~\ref{fig:reading_giving}, interpretation is inherently
subjective, while explanation is fundamentally relational and communicative.

This distinction provides a conceptual basis for separating interpretation and
explanation in AI systems, and directly motivates the Interpret/Explain Schema
introduced in Section~\ref{sec:inter}.

\subsection{Explanation: Why?}\label{sec2.4}

The motivations for explanation in AI have been extensively discussed in the
literature \citep{lipton2018mythos, arrieta2020explainable, hall2019introduction}.
These motivations can be broadly grouped into commercial, regulatory, and
technical considerations \citep{zhong2021ai}.

While these drivers explain why explanations are desirable, they do not, by
themselves, specify how explanations should be structured, nor how different
stakeholders' needs should be aligned within an AI system.
As such, they motivate the need for explanation frameworks, but do not resolve
the conceptual and structural ambiguities identified above.

\subsection{Explanation: How?}\label{sec2.5}

Existing work on explanation methods is commonly organised around a small set of
taxonomic distinctions \citep{lipton2018mythos, guidotti2018survey}.
A first distinction separates \emph{model-intrinsic} approaches, which rely on
transparent model structures, from \emph{post-hoc} approaches, which extract
explanatory information from otherwise opaque models.

Post-hoc methods are further categorised along axes such as local vs.\ global
scope, and model-specific vs.\ model-agnostic applicability.
These distinctions are useful for comparing techniques; richer frameworks
also encode mechanisms, model access, stakeholders, and interaction
\citep{schwalbe2024taxonomy,wang2024roadmap}. Their published categories do
not, however, require a single record that links each evidence source to its
transformation and to the analyst--recipient transition.

Hybrid approaches---such as attribution-guided counterfactuals or
concept-based models combined with post-hoc analysis---can carry several
valid taxonomic labels. The question here is how to record their ordered
operations and assumptions alongside those labels.
This limitation motivates the need for a framework that captures explanation
generation as a structured process rather than as a static category.

\subsection{Explanation: How to Evaluate?}\label{sec:2.6}

Evaluation strategies for explanations in AI are commonly classified into
application-grounded, human-grounded, and functionally-grounded approaches
\citep{doshi2017towards}.
These strategies address different aspects of explanation quality, including
usefulness, human understanding, and fidelity to the underlying model.

Importantly, these evaluation frameworks assess \emph{how good} an explanation
is, but do not specify \emph{how explanations are structurally generated}.
This distinction is central to this paper: the framework proposed in the
following sections focuses on the structure and expressiveness of explanation
pipelines, while remaining compatible with existing evaluation methodologies.

\subsection{Related Surveys and the Role of GFGE}\label{sec:related_surveys}

Recent reviews give increasingly precise ways to organise XAI, and thus
provide an important test of the contribution claimed here. Schwalbe and
Finzel's survey of surveys offers a detailed taxonomy of method properties
\citep{schwalbe2024taxonomy}. Wang et al. connect stakeholder questions to
methods across an AI system's lifecycle \citep{wang2024roadmap}. More recent
taxonomies support developers' technique selection or distinguish mechanisms
and levels of interpretability \citep{cousineau2025taxonomy,elsayed2026taxonomy}.
Human-centred reviews instead organise design goals and evaluation evidence
\citep{mangold2026human}; the LLM-specific review by Cai et al. concentrates
on whether interpretability enables interventions and model improvement
\citep{cai2026actionable}. Table~\ref{tab:survey_comparison} compares these
contributions with the question GFGE addresses.

\begin{table}[t]
\centering\scriptsize
\caption{Scope of related XAI reviews and the complementary role of GFGE.}
\label{tab:survey_comparison}
\begin{tabular}{@{}>{\raggedright\arraybackslash}p{2.55cm}>{\raggedright\arraybackslash}p{3.8cm}>{\raggedright\arraybackslash}p{4.2cm}@{}}
\toprule
\textbf{Review} & \textbf{Primary organising question} & \textbf{GFGE's distinct analytical question} \\
\midrule
Schwalbe and Finzel~\citep{schwalbe2024taxonomy}
& Which properties distinguish methods and guide their selection?
& Which interpretative operations and information flows generate a particular explanation? \\
\midrule
Wang et al.~\citep{wang2024roadmap}
& Which stakeholders need which explanations at which lifecycle stage?
& How does a selected method transform data, model, and output evidence into an account for that stakeholder? \\
\midrule
Cousineau and Dara~\citep{cousineau2025taxonomy}; Elsayed et al.~\citep{elsayed2026taxonomy}
& How should techniques be classified or selected by mechanism and interpretability level?
& How are intrinsic, post-hoc, and hybrid techniques instantiated in the same modular procedure? \\
\midrule
Mangold et al.~\citep{mangold2026human}
& How should XAI systems be designed and evaluated with human users?
& Which generation steps precede that evaluation, and where does recipient interpretation enter? \\
\midrule
Cai et al.~\citep{cai2026actionable}
& How can LLM interpretability support intervention and model improvement?
& How do internal evidence, intervention tests, and communicated explanations fit into a framework spanning model families? \\
\bottomrule
\end{tabular}
\end{table}

The comparison does not imply that taxonomies omit stakeholders or that GFGE
replaces user-centred evaluation. Rather, the proposed contribution is an
explicit procedural mapping grounded in the IES: a method is described by the
roles it instantiates, the evidence it transfers between them, and the
transition from analyst sense-reading to recipient-oriented sense-giving.

\subsection{Summary and Identified Gap}\label{sec:2.7}

The literature reviewed in this section shows that explainable AI has been
studied extensively, yet through several lines of work that remain only weakly
connected.
In particular, existing works provide complementary contributions:

\begin{itemize}
  \item \textbf{Taxonomies of XAI methods}, which organise techniques along axes
  such as intrinsic vs.\ post-hoc, local vs.\ global, or model-specific vs.\
  model-agnostic \citep{arrieta2020explainable, molnar2020interpretable,
  guidotti2018survey};

  \item \textbf{Informal or discipline-specific definitions} of interpretability
  and explainability, typically grounded in philosophical, psychological, or
  technical intuitions \citep{miller2019explanation, lipton2018mythos,
  doshi2017towards};

  \item \textbf{Evaluation frameworks}, which propose strategies to assess
  explanation quality (e.g., application-grounded, human-grounded, and
  functionally-grounded evaluation)
  \citep{doshi2017towards, mohseni2021multidisciplinary, hoffman2018metrics}.
\end{itemize}

These literatures already model method mechanisms, stakeholders, evaluation,
and interaction in substantial detail. The narrower gap pursued here is a
common representation that jointly records evidence provenance, the order of
interpretative operations, and the transition from an analyst's findings to
a recipient-facing account. Hybrid procedures make the need for such a
record especially visible.

The reviewed approaches address parts of this problem. Their emphases differ
from the combination developed here:

\begin{enumerate}
  \item \textbf{Working definitions} of interpretability and
  explainability grounded in a distinction between
  \emph{sense-reading} and \emph{sense-giving};

  \item \textbf{A system-level schema} that explicitly models the roles of
  different stakeholders (e.g., developers vs.\ end-users) and links their
  interpretations of data, models, and outputs within the same AI pipeline;

  \item \textbf{A general procedural framework} that represents explanation
  generation as a structured workflow and can encode heterogeneous explanation
  families---including intrinsic, post-hoc, and hybrid approaches---within a single,
  modular structure.
\end{enumerate}

GFGE's target is the link between these levels: \emph{how interpretation and
explanation jointly arise in an AI system} and how methods with different
technical mechanisms can be expressed within one procedure across
stakeholders. A mapping at this level complements method selection and
evaluation; it does not establish that a mapped explanation is faithful or
understood.
Addressing this gap is the objective of the next section, which introduces
general definitions of interpretability and explainability, followed by the
Interpret/Explain Schema (IES) and the General Framework for Generating
Explanations (GFGE).

\section{Interpretation and Explanation: Conceptual Basis}\label{EIS}

Explanation in AI draws on several traditions. Causal accounts ask what
brought about an outcome, whereas social and cognitive accounts ask how an
account answers a person's question in context
\citep{lewis1986causal,josephson1996abductive,malle2006mind,
lombrozo2006structure,miller2019explanation}. Both matter here: a claim
about model behaviour may require causal evidence, while its communication
must still be intelligible to its intended recipient.

Knowledge-management work distinguishes \emph{sense-reading}, the
interpretation of received information, from \emph{sense-giving}, the
assignment of meaning to information conveyed to others
\citep{despres1999knowledge,arduin2015information}. Individuals read
information through interpretative frameworks \citep{jones2011mental} or
mental models \citep{tsuchiya1993improving}. Consequently, two people may
derive different meanings from the same display. Figure~\ref{fig:reading_giving}
introduces this interaction; it is the conceptual basis of the IES and GFGE.

\begin{definition}\label{definition_interpret}
\emph{Interpretability} is the degree to which an observer can derive a
meaningful understanding from an event, representation, model, or account.
\end{definition}

\begin{definition}\label{definition_explain}
\emph{Explainability} is the capacity to construct and communicate an account
intended to make an event or outcome understandable, including its reasons
or causes when those claims are warranted.
\end{definition}

These definitions assign different roles to the acts, not disjoint labels to
AI systems. An analyst may interpret data or model behaviour before producing
an explanation; a recipient subsequently interprets that explanation.
Sense-giving does not guarantee successful sense-reading. Likewise, an
account of association is not automatically a causal account. The IES
locates these acts in an AI system, and GFGE identifies the operations that
can connect them.

These are conceptual, observer-relative working definitions rather than
measurement scales. In particular, ``meaningful'' and ``understandable'' depend
on a person's task and background knowledge. The names of GFGE's
\texttt{Interpret...} modules denote computational extraction or organisation
of \emph{candidate evidence}. Running such a module is not itself a person's
cognitive interpretation. Whether the output becomes meaningful to an analyst
or recipient requires separate observation and evaluation.

\section{Interpret/Explain Schema for AI Systems} \label{sec:inter}

Building on the definitions established in Section~\ref{EIS}, this section examines how
interpretation and explanation materialise within an AI system. By analysing the flow of information between developers, models, data, and users, we introduce an Interpret/Explain Schema (IES) that characterises the roles of sense-reading and
sense-giving in computational settings. This schema then motivates the design of a General Framework for Generating Explanations (GFGE), presented in Section~\ref{sec4.3.4}, which provides a model-agnostic representation of explanation workflows across AI applications.

\subsection{AI: To Interpret or to Explain?}

\begin{figure}[t]
\centering
\begingroup
\input{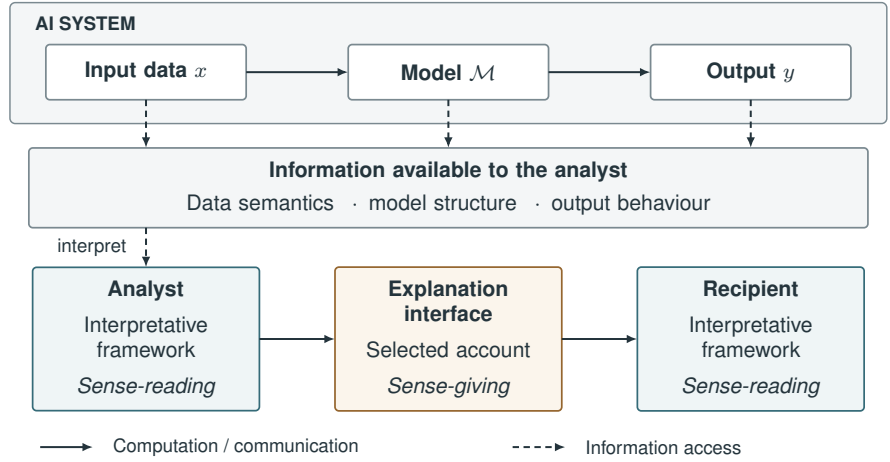}
\begin{tikzpicture}[gf base]
  \draw[draw=gfRule,line width=0.55pt,fill=gfSystemFill,rounded corners=2pt]
    (0,3.4) rectangle (11.6,5.0);
  \node[anchor=west,gf small,font=\fontfamily{phv}\selectfont\fontsize{7}{8.5}\selectfont\bfseries]
    at (0.22,4.72) {AI SYSTEM};
  \node[gf box,text width=2.25cm,minimum width=2.65cm,minimum height=0.73cm]
    (data) at (1.8,4.08) {\textbf{Input data} $x$};
  \node[gf box,text width=2.25cm,minimum width=2.65cm,minimum height=0.73cm]
    (model) at (5.8,4.08) {\textbf{Model} $\mathcal M$};
  \node[gf box,text width=2.25cm,minimum width=2.65cm,minimum height=0.73cm]
    (output) at (9.8,4.08) {\textbf{Output} $y$};
  \draw[gf flow] (data.east) -- (model.west);
  \draw[gf flow] (model.east) -- (output.west);

  \node[gf box,fill=gfSystemFill,text width=10.7cm,
    minimum width=11.1cm,minimum height=0.78cm]
    (information) at (5.8,2.55)
    {\textbf{Information available to the analyst}\\[3pt]
      Data semantics \enspace$\cdot$\enspace model structure
      \enspace$\cdot$\enspace output behaviour};
  \draw[gf access] (data.south) -- (data.south |- information.north);
  \draw[gf access] (model.south) -- (model.south |- information.north);
  \draw[gf access] (output.south) -- (output.south |- information.north);

  \node[gf human,text width=2.6cm,minimum width=3cm,minimum height=1.8cm]
    (analyst) at (1.8,0.55)
    {\textbf{Analyst}\\[4pt]Interpretative framework\\[4pt]
      \textit{Sense-reading}};
  \node[gf account,text width=2.6cm,minimum width=3cm,minimum height=1.8cm]
    (interface) at (5.8,0.55)
    {\textbf{Explanation interface}\\[4pt]Selected account\\[4pt]
      \textit{Sense-giving}};
  \node[gf human,text width=2.6cm,minimum width=3cm,minimum height=1.8cm]
    (recipient) at (9.8,0.55)
    {\textbf{Recipient}\\[4pt]Interpretative framework\\[4pt]
      \textit{Sense-reading}};
  \draw[gf access] (analyst.north |- information.south) --
    node[left=4pt,gf small]{interpret} (analyst.north);
  \draw[gf flow] (analyst.east) -- (interface.west);
  \draw[gf flow] (interface.east) -- (recipient.west);

  \draw[gf flow] (0.4,-0.88) -- (1.1,-0.88);
  \node[anchor=west,gf small] at (1.25,-0.88)
    {Computation / communication};
  \draw[gf access] (6.65,-0.88) -- (7.35,-0.88);
  \node[anchor=west,gf small] at (7.5,-0.88)
    {Information access};
\end{tikzpicture}
\endgroup
\caption{Interpretation and explanation in an AI system. The analyst reads
information from data, model, and output; an interface conveys an account to
the recipient, who interprets it through their own mental model. Solid arrows
denote computation or communication; dashed arrows denote information access.}
\label{fig:Interpret_Explain_scheme}
\end{figure}

An AI system connects input data, a model, and its output, as shown in
Figure~\ref{fig:Interpret_Explain_scheme}. An analyst---for example, a
developer, domain specialist, or auditor---interprets information available
from each component through an interpretative framework. This is
\emph{sense-reading} \citep{jones2011mental,tsuchiya1993improving}. Selecting
and presenting some of that information through an interface is
\emph{sense-giving}; a recipient then interprets the account through their own
mental model \citep{arduin2015information,miller2019explanation}. The IES thus
locates explanation in the interaction between evidence, its presentation,
and recipient interpretation. The recipient's understanding is an empirical
question, not an automatic consequence of making model information visible.

\subsection{Definitions in the Context of AI}\label{sec:definitions_AI}

The IES distinguishes what information an analyst can inspect from what a
recipient can understand. Applying Definitions~\ref{definition_interpret}
and~\ref{definition_explain} yields five system-level capacities.

\begin{definition}\label{def:interpretability_of_data}
\emph{Interpretability of data} is the degree to which an observer can assign
meaning to input features, their measurements, and their constraints.
\end{definition}

\begin{definition}\label{def:interpretability_of_model}
\emph{Interpretability of a model} is the degree to which an observer can
understand the model structures or mechanisms relevant to a question.
\end{definition}

\begin{definition}\label{def:explainability_of_results}
\emph{Explainability of a result} is the capacity to construct an account of
a model output for a specified question and audience.
\end{definition}

\begin{definition}\label{def:post_hoc_interpretability}
\emph{Post-hoc interpretability} is the capacity to derive information about
a trained model or its behaviour using an additional analytic method.
\end{definition}

\begin{definition}\label{def:interpretability_of_explanations}
\emph{Interpretability of an explanation} is the degree to which a recipient
can understand the communicated account in the relevant context.
\end{definition}

The first, second, and fourth capacities primarily concern evidence
available to the analyst; the fifth concerns a recipient's sense-reading.
The third connects them through sense-giving. A post-hoc method may be useful
even when internal model structure is available, and an intrinsically
interpretable model still needs its relevant result communicated. These
definitions describe capacities, not guarantees of fidelity, causal validity,
or user benefit. Those properties require separate evaluation
\citep{doshi2017towards,piculin2025userstudies}.

\subsection{Discussion: Stakeholders and Explanatory Needs}

A developer may inspect data quality or model structure, a domain expert may
ask whether a prediction accords with domain knowledge, and an affected user
may ask why an outcome occurred or what could change it. The same system can
therefore support several explanation questions and audiences. The IES marks
the point at which the analyst's interpretation becomes an account offered to
another person; GFGE will make the operations behind that point explicit.

The reasons for explaining include debugging, accountability, appropriate
reliance, and decision support \citep{arrieta2020explainable,zhong2021ai}.
Whether an intrinsic model or a post-hoc method is suitable depends on the
question, the available evidence, and the consequences of error
\citep{rudin2019stop,carvalho2019machine}. Neither a transparent model nor a
plausible post-hoc artefact automatically satisfies the recipient's need.
This motivates a framework that can represent both sources of evidence and
the communication step in one procedure.

\subsection{General Framework for Generating Explanations}
\label{sec4.3.4}

Building upon the Interpret/Explain Schema (IES) introduced in
Figure~\ref{fig:Interpret_Explain_scheme} and the definitions formalised in
Section~\ref{sec:definitions_AI}, we now propose a General Framework for
Generating Explanations (GFGE). The purpose of GFGE is to operationalise the
IES by providing a modular representation of explanation procedures in AI
systems. In line with the IES, an analyst interprets information from data,
model structure, and outputs, then selects findings relevant to a recipient's
question \citep{doshi2017towards,miller2019explanation}.

\paragraph{Novelty and Motivation.}
GFGE specifies roles rather than prescribing a model family or explainer.
Its modules can therefore be instantiated by a rule list, a tree ensemble,
a neural model, or a combination of intrinsic and post-hoc operations.
This modularity is the basis for the unification claim developed in
Section~\ref{sec:instantiation}; the actual evidence available still depends
on the model, data, and method.

\paragraph{Inputs of the Framework.}
GFGE takes as input the following components, each corresponding to specific
forms of sense-reading or sense-giving as depicted in
Figure~\ref{fig:reading_giving}:
\begin{itemize}
    \item $\mathcal{M}$: the model in deployment.
    \item $\mathcal{D}$: the input instance or collection whose model behaviour is
    to be explained.
    \item $\mathcal{P}$: a preprocessing function that prepares the data for $\mathcal{M}$.
    \item $q$, $u$, and $c$: the question, audience, and context. Context
    includes permitted changes and domain assumptions.
    \item \texttt{InterpretData}: interprets information from $\mathcal{D}$, corresponding to
    the interpretability of data in Definition~\ref{def:interpretability_of_data}.
    \texttt{InterpretData} denotes the data-facing evidential role; domain or
    feasibility constraints supplied through $c$ may originate from external
    knowledge and are not assumed to be learned from $\mathcal{D}$.
    \item \texttt{InterpretModel}: interprets information from $\mathcal{M}$, corresponding to
    model interpretability in Definition~\ref{def:interpretability_of_model}.
    \item \texttt{InterpretOutput}: extracts evidence about observed or queried
    outputs of $\mathcal{M}$; it does not define the desired target in $q$.
    \item \texttt{PostHocInterpretation}: when needed, extracts additional evidence
    from model behaviour, for example by fitting a LIME surrogate
    \citep{ribeiro2016should} or computing SHAP attributions
    \citep{lundberg2017unified}. It is optional for an intrinsically
    interpretable model.
    \item \texttt{Aggregation}: selects and communicates already recorded claims for
    $q$ and $u$, without silently adding new evidential claims.
\end{itemize}

\paragraph{Evidence Records and Role Assignment.}
Let $q=(t,\tau,\kappa,\gamma)$ identify question type
$t\in T_q$, target $\tau$, contrast $\kappa$, and requested constraints
$\gamma$, where $T_q$ may include why, why-not, what-if, how-to, and
global-pattern questions; these examples are not exhaustive.
A proposed outcome is part of $q$, whereas observing
whether the model produces it is an output operation. Let
$r_i=(s_i,p_i,k_i,\pi_i,\sigma_i,a_i,\ell_i)$ record each candidate claim $k_i$,
its source $s_i$, producing operation $p_i$, parent-record identifiers $\pi_i$,
scope $\sigma_i$, assumptions $a_i$, and limitations $\ell_i$. The parent
identifiers expose which earlier claims a later analysis consumes. The first
four modules return sets of such records; $A$ returns a selected account $e$.
The communicated claims in $e$ retain links to the corresponding records
in $r$.
Table~\ref{tab:module_contracts} is the coding contract used below.

\begin{table}[t]
\centering\scriptsize
\caption{GFGE module contracts. $E_0=E_D\cup E_M\cup E_O$ and
$E=E_0\cup E_H$ denote available records; $q$ specifies the target.}
\label{tab:module_contracts}
\begin{tabular}{@{}>{\raggedright\arraybackslash}p{1.15cm}>{\raggedright\arraybackslash}p{2.5cm}>{\raggedright\arraybackslash}p{3.35cm}>{\raggedright\arraybackslash}p{3.45cm}@{}}
\toprule
\textbf{Role} & \textbf{Input $\to$ output} & \textbf{Necessary assignment condition} & \textbf{Exclude and assign elsewhere} \\
\midrule
ID & $(\mathcal D,z,c,q)\to E_D$ & Extracts feature meaning, measurement, domain constraints, or feasible changes. & Desired outcome belongs to $q$; a model response belongs to IO. \\
IM & $(\mathcal M,z,q)\to E_M$ & Directly reads a mechanism, parameter, rule, leaf, prototype, or circuit actually used by the predictor. & A fitted surrogate or attribution computation belongs to PH, even if model internals accelerate it. \\
IO & $(y,\mathcal M,z,q)\to E_O$ & Records observed predictions, scores, or queried contrasts produced by $\mathcal M$. & The requested alternative is in $q$; searching for inputs that attain it belongs to PH. \\
PH & $(\mathcal M,z,y,q,E_0)\to E_H$ & Runs an additional explanatory analysis: perturbation, attribution, surrogate fitting, intervention test, or candidate search. & Simply reading an existing model rule is IM; wording an existing claim is A. \\
A & $(E,q,u)\to e$ & Selects, orders, and renders supported claims with provenance for audience $u$. & Any new model query, statistical estimate, or factual inference must be recorded in IO, PH, ID, or IM first. \\
\bottomrule
\end{tabular}
\end{table}

These are \emph{operation} roles, not mutually exclusive method labels. Split a
multi-step method at the point where provenance changes; for example, reading
forest leaves is IM, searching their regions is PH, and checking each
candidate prediction is IO. A role is inactive only if no claim needed to
answer $q$ requires it; a missing necessary source is marked \emph{unresolved},
not silently treated as inactive. The analyst may interpret these records;
computing them does not establish human understanding. Audience $u$ enters
selection and rendering in $A$, while evidence extraction is held fixed for a
given $q$ and $c$. A recipient's follow-up can change $q$ or $c$ and trigger
a new pass. This choice makes audience dependence explicit and consistent.

\paragraph{The Framework.}
Figure~\ref{fig:gfge_diagram} depicts the information flow from the IES to
GFGE: the three system sources feed interpretation modules, optional post-hoc
analysis contributes further evidence, and aggregation selects an account for
the question and audience. Algorithm~\ref{alg:gfge} illustrates the common
single-pass case and is not the general semantics of GFGE. The general
representation is the operation graph in Equation~\ref{eq:gfge_mapping},
whose edges determine method-specific dependencies; a valid graph may place
PH before IM or contain repeated roles.

\begin{figure}[p]
\centering
\begingroup
\input{Figures/journal_style_v5.tex}
\begin{tikzpicture}[gf base]
  \node[gf heading] at (0,9.9)
    {\textbf{a}\quad Evidence, communication and recipient feedback};
  \node[gf box,fill=gfSystemFill,text width=8.35cm,
    minimum width=8.8cm,minimum height=0.88cm] (system) at (4.4,8.95)
    {\textbf{AI system}\\[3pt]
      Input $x$ \enspace$\xrightarrow{\mathcal P}$\enspace $z$
      \enspace$\xrightarrow{\mathcal M}$\enspace Output $y$};
  \node[gf box,text width=2cm,minimum width=2.35cm,
    minimum height=0.88cm] (request) at (10.4,8.95)
    {\textbf{Question $q$}\\Context $c$};

  \draw[draw=gfTeal,line width=0.65pt,fill=gfTealFill,rounded corners=2pt]
    (0,4.78) rectangle (8.8,7.95);
  \node[gf heading] at (0.22,7.59)
    {Roles used to label operations in $G_f$};
  \node[gf small,anchor=west] at (0.22,7.20)
    {Roles may repeat; dependencies are method-specific (panel b).};
  \node[gf box,text width=3.42cm,minimum width=3.9cm,
    minimum height=0.86cm] (id) at (2.2,6.51)
    {\textbf{ID\quad Data}\\Semantics and constraints};
  \node[gf box,text width=3.42cm,minimum width=3.9cm,
    minimum height=0.86cm] (im) at (6.6,6.51)
    {\textbf{IM\quad Model}\\Predictor mechanisms};
  \node[gf box,text width=3.42cm,minimum width=3.9cm,
    minimum height=0.86cm] (io) at (2.2,5.48)
    {\textbf{IO\quad Output}\\Actual model responses};
  \node[gf box,text width=3.42cm,minimum width=3.9cm,
    minimum height=0.86cm] (ph) at (6.6,5.48)
    {\textbf{PH\quad Post-hoc}\\Additional analysis};
  \draw[gf flow] (system.south) -- (4.4,7.95);
  \draw[gf access,rounded corners=2pt] (request.west) -- (9.04,8.95)
    -- (9.04,7.53) -- (8.8,7.53);

  \node[gf box,fill=gfSystemFill,text width=8.35cm,
    minimum width=8.8cm,minimum height=1.32cm] (record) at (4.4,3.70)
    {\textbf{Evidence record $r$}\quad\textit{Unresolved gaps retained}\\[3pt]
      Source $s$ $\cdot$ operation $p$ $\cdot$ claim $k$ $\cdot$ parent records $\pi$\\[2pt]
      Scope $\sigma$ $\cdot$ assumptions $a$ $\cdot$ limitations $\ell$};
  \draw[gf flow] (4.4,4.78) -- (record.north);

  \node[gf account,text width=8.35cm,minimum width=8.8cm,
    minimum height=0.95cm] (aggregate) at (4.4,2.09)
    {\textbf{A\quad Aggregation}\\[3pt]
      Select for $q,u$ and communicate supported claims\quad\textit{Sense-giving}};
  \draw[gf flow] (record.south) -- (aggregate.north);
  \node[gf human,text width=2cm,minimum width=2.35cm,
    minimum height=0.95cm] (audience) at (10.4,2.09)
    {\textbf{Audience $u$}\\Needs and background};
  \draw[gf access] (audience.west) -- (aggregate.east);

  \node[gf account,text width=3.42cm,minimum width=3.9cm,
    minimum height=0.95cm] (explanation) at (2.2,0.48)
    {\textbf{Explanation artefact $e$}\\Selected, traceable claims};
  \node[gf human,text width=3.42cm,minimum width=3.9cm,
    minimum height=0.95cm] (recipient) at (6.6,0.48)
    {\textbf{Recipient}\\\textit{Sense-reading}};
  \draw[gf flow] (2.2,1.615) -- (explanation.north);
  \draw[gf flow] (explanation.east) -- (recipient.west);
  \draw[gf access,rounded corners=3pt] (recipient.east) -- (12,0.48)
    -- (12,8.95) -- (request.east);
  \node[gf small,align=center,text width=2.35cm] at (10.48,6.02)
    {Recipient feedback $f_t$\\[3pt]updates $q,c$};

  \draw[gf flow] (0.15,-0.55) -- (0.9,-0.55);
  \node[gf small,anchor=west] at (1.05,-0.55)
    {Evidence / communication};
  \draw[gf access] (5.65,-0.55) -- (6.4,-0.55);
  \node[gf small,anchor=west] at (6.55,-0.55)
    {Context / audience / feedback};

  \draw[draw=gfRule,line width=0.55pt] (0,-1.10) -- (12,-1.10);
  \node[gf heading] at (0,-1.58)
    {\textbf{b}\quad Selected dependency paths};
  \node[gf small,anchor=east] at (12,-1.58)
    {Partial paths; other roles omitted};

  \node[align=left,anchor=west,text width=2.2cm] at (0,-2.50)
    {\textbf{CBM-zero}};
  \node[gf box,text width=3.07cm,minimum width=3.45cm,
    minimum height=0.92cm] (cbmph) at (4.1,-2.50)
    {\textbf{PH}\\Construct concept map};
  \node[gf box,text width=3.85cm,minimum width=4.23cm,
    minimum height=0.92cm] (cbmim) at (9.15,-2.50)
    {\textbf{IM}\\Inspect converted predictor};
  \draw[gf flow] (cbmph.east) -- (cbmim.west);

  \node[align=left,anchor=west,text width=2.2cm] at (0,-3.90)
    {\textbf{SHAP-guided}\\\textbf{counterfactual}};
  \node[gf box,text width=2.08cm,minimum width=2.46cm,
    minimum height=0.92cm] (attrib) at (3.65,-3.90)
    {\textbf{PH}\\Attribution};
  \node[gf box,text width=2.08cm,minimum width=2.46cm,
    minimum height=0.92cm] (search) at (6.83,-3.90)
    {\textbf{PH}\\Search};
  \node[gf box,text width=2.08cm,minimum width=2.46cm,
    minimum height=0.92cm] (response) at (10.02,-3.90)
    {\textbf{IO}\\Responses};
  \draw[gf flow] (attrib.east) -- (search.west);
  \draw[gf flow] (search.east) -- (response.west);
\end{tikzpicture}
\endgroup
\caption{GFGE roles, evidence, and communication. \textbf{(a)} Computational
operations supply evidence for analyst sense-reading; aggregation performs
sense-giving, followed by recipient sense-reading. The role panel lists the
vocabulary of $G_f$, without imposing an execution order. Solid arrows denote
evidence or communication; dashed arrows denote contextual inputs or feedback.
\textbf{(b)} Selected dependency paths show PH before IM in CBM-zero and
repeated PH operations in attribution-guided counterfactuals
\citep{wang2025cbmzero,zhong2022shap}. These paths omit other roles.}
\label{fig:gfge_diagram}
\end{figure}

\begin{algorithm}[t]
\caption{Canonical single-pass instantiation of GFGE}\label{alg:gfge}
\DontPrintSemicolon
\KwIn{Model $\mathcal{M}$; data $\mathcal{D}$; preprocessing $\mathcal{P}$;
question $q$; audience $u$; context $c$; implemented modules $I_D,I_M,I_O,H,A$}
\KwOut{Explanation artefact $e$ and an evidence record}
$z \leftarrow \mathcal{P}(\mathcal{D})$; $y \leftarrow \mathcal{M}(z)$\;
$d \leftarrow I_D(\mathcal{D},z,c,q)$\;
$m \leftarrow I_M(\mathcal{M},z,q)$\;
$o \leftarrow I_O(y,\mathcal{M},z,q)$\;
$h \leftarrow \varnothing$\;
\If{additional post-hoc evidence is needed}{
  $h \leftarrow H(\mathcal{M},z,y,q,d,m,o)$\;
}
$E \leftarrow d\cup m\cup o\cup h$\;
$r \leftarrow \operatorname{Record}(E)$\;
$e \leftarrow A(E,q,u)$\;
\Return $(e,r)$
\end{algorithm}
\FloatBarrier

\begin{samepage}
The present formulation treats a model interaction or generated trace as an
output $y$. Richer sequential or agentic systems may require repeated GFGE
passes or an extension of the output representation.\label{par:gfge_scope}
\par
\end{samepage}

Here $I_D$, $I_M$, $I_O$, $H$, and $A$ instantiate
\texttt{InterpretData}, \texttt{InterpretModel}, \texttt{InterpretOutput},
\texttt{PostHocInterpretation}, and \texttt{Aggregation}, respectively. The record $r$
retains the source, scope, assumptions, and limitations of the findings used
to construct $e$. This is a workflow template: it does not compute an
explanation until a method supplies the modules, their data access, and
their decision rules. The displayed order is illustrative only;
method-specific dependency order is represented by $G_f$ in
Equation~\ref{eq:gfge_mapping}. A recipient can supply feedback $f_t$ after round $t$;
an interaction policy then updates $(q_{t+1},c_{t+1})=
U(q_t,c_t,f_t)$ and reruns the relevant modules. The policy $U$ is
application-specific, and recipient understanding remains an empirical matter.

\paragraph{Relation to Existing Explanation Taxonomies.}
GFGE provides a unifying lens through which standard distinctions can be
reinterpreted:
\begin{itemize}
    \item \textbf{Model-intrinsic vs post-hoc.}  
    An intrinsic method can expose prediction-relevant structure in IM and
    outcomes in IO. A post-hoc method records its additional analysis in PH;
    this remains true when model internals are also available.
    \item \textbf{Local vs global.}
    Scope is determined by the explanatory question and evidence examined.
    LIME is typically local, whereas aggregating local attributions across a
    dataset can support a global analysis \citep{ribeiro2016should,lundberg2017unified}.
    \item \textbf{Model-specific vs model-agnostic.}
    This distinction concerns module implementation. TreeSHAP exploits tree
    structure, while a model-agnostic explainer can use input--output queries
    \citep{lundberg2018consistent,ribeiro2016should}.
\end{itemize}

\paragraph{Summary.}
GFGE is an operational instantiation of the IES: it encodes the schema's
information sources and stakeholder transition as modular roles. Its
generality lies in allowing those roles to be populated by different
techniques, from intrinsic model inspection to post-hoc querying. The mapping
in Section~\ref{sec:instantiation} examines the scope of that claim.

\paragraph{Structural Expressiveness.}
For a method $f$ and a question--audience--context tuple $(q,u,c)$, define
its GFGE representation as a role-typed operation graph and evidence record:
\begin{equation}
\Phi(f;q,u,c)=(G_f,r),\qquad
G_f=(V_f,E_f),\quad
\lambda:V_f\to\{ID,IM,IO,PH,A\}.
\label{eq:gfge_mapping}
\end{equation}
Here, \emph{role-typed} denotes role labels constrained by assignment rules;
it does not imply a formal type system for checking composition.
Each node is an operation assigned one role by the contracts in
Table~\ref{tab:module_contracts}; an edge $(v_i,v_j)$ means that $v_j$
consumes evidence from $v_i$. The parent identifiers $\pi_i$ make this
dependency explicit at claim level. A method may repeat a role in multiple
nodes, and a role may be inactive under the rule above. GFGE represents a procedure \emph{within its stated
scope} only when every claim required by the declared evidential requirements
for $q$ and $c$ can be assigned a source and
operation, dependencies can be ordered, and $A$ can link communicated claims
back to evidence. These requirements must be specified for the application;
GFGE checks their procedural coverage without providing an independent
criterion for epistemic or semantic completeness.
A mapping fails or remains partial if, for example, a
human's unstated domain judgment supplies a claim without a recorded source,
an interactive method changes goals without a representable context update,
or the only available artefact is an untraceable generated rationale. These
cases are recorded as unresolved rather than forced into PH or A. The
structural scope claim can be challenged by methods whose required operations
cannot be assigned to the defined roles; low agreement among independent
coders would instead challenge the operational reproducibility of the coding
contract. Structural fit does not establish correctness, fidelity,
understanding, or usefulness.

\paragraph{Relation to Evaluation and Explanation Quality.}
By design, GFGE does not define or optimise explanation quality.
Quality criteria such as fidelity, usefulness, trustworthiness, actionability,
or human understanding depend on the target audience and application context.
These aspects are addressed through evaluation strategies discussed in
Section~\ref{sec:2.6}, including application-grounded, human-grounded, and
functionally-grounded evaluation.
GFGE provides the structural substrate upon which such evaluations can be
systematically applied, by clarifying which components of the explanation
pipeline are responsible for which explanatory roles.

\subsection{Discussion}
\paragraph{The XAI-Specific Contribution of GFGE.}
GFGE specialises provenance, dataflow, workflow, and evidence representations
through five joint
commitments: (i) an XAI role vocabulary that distinguishes data semantics,
predictive mechanisms, observed outputs, additional explanatory analysis, and
communication; (ii) the explanatory question $q$, recipient $u$, and context
$c$ that determine which claims an account must support; (iii) the IES
transition from analyst sense-reading to recipient-oriented sense-giving and
subsequent recipient sense-reading; (iv) provenance, assumptions, and
limitations attached to individual explanatory claims; and (v) explicit
records of unresolved explanatory dependencies. Together, these commitments
connect the flow of computational evidence to the conditions under which it
can support a particular explanatory account. For example, a counterfactual
may have fully traceable search provenance while lacking the feasibility
evidence needed to answer a how-to question. GFGE makes that missing
explanatory dependency part of the representation. Its contribution is this
common representational contract for XAI workflows; the graph structure
itself is not claimed as a new formalism.

\paragraph{What GFGE Enables Beyond Existing Taxonomies.}
XAI surveys and taxonomies (e.g., intrinsic vs.\ post-hoc,
local vs.\ global, model-specific vs.\ model-agnostic) provide valuable
\emph{descriptive axes} for categorising explanation methods.
These axes and newer mechanism-based taxonomies
\citep{schwalbe2024taxonomy,elsayed2026taxonomy} describe important method
properties, but a category alone does not specify the full \emph{procedural
workflow} through which evidence becomes an account for a stakeholder.

The Interpret/Explain Schema (IES) and the General Framework for Generating
Explanations (GFGE) go beyond such classifications by explicitly modelling
explanations as \emph{stakeholder-mediated pipelines} grounded in
sense-reading and sense-giving.
In particular, GFGE specifies a \emph{procedural contract}: an analysis of an
explanation method should state (i) how data are rendered meaningful
(\texttt{InterpretData}), (ii) what information is available from the model
(\texttt{InterpretModel}), (iii) how outputs are interpreted
(\texttt{InterpretOutput}), (iv) whether post-hoc interpretation is required, and
(v) how these elements are aggregated into a communicative explanation for
the intended recipient (\texttt{Aggregation}). A role can be inactive, but
that decision should be explicit.

This procedural view supports analyses that a taxonomic label alone does not
specify.
First, GFGE supports \emph{procedural completeness checks} relative to a
specified question $q$, context $c$, and declared evidential requirements,
as well as \emph{failure diagnosis}: explanation methods
can be analysed to identify missing or weak components (e.g., absent semantic
mapping or aggregation), even when they are correctly labelled as post-hoc or
intrinsic.
Second, GFGE naturally represents \emph{hybrid explanation pipelines}---such as
attribution-guided counterfactuals or concept-based models with post-hoc
influence analysis---without forcing them into ambiguous or overlapping
categories.
Third, by explicitly distinguishing analyst sense-reading from
recipient-oriented sense-giving, GFGE makes the transition between evidence
and communication inspectable; stakeholder roadmaps identify the audience but
do not necessarily trace this transition within each method
\citep{wang2024roadmap}.

Recent methods illustrate the utility of this distinction. ReCalX shows that
perturbation-based attributions depend on how the model behaves under the
queries used by the post-hoc module: unreliable probabilities at perturbed
inputs weaken the evidence before aggregation begins \citep{decker2025recalx}.
EECE uses random-forest leaf structure to generate candidate counterfactuals;
GFGE therefore records both model-structure access in
\texttt{InterpretModel} and alternative-outcome search in
\texttt{PostHocInterpretation}, while data constraints and the communicated
contrast remain separate roles \citep{zhang2025efficient}. Causal
backtracking counterfactuals make the assumptions on feasible interventions
particularly visible in \texttt{InterpretData} \citep{fatemi2025backtracking}.

Concept methods and language-model explanations expose a different boundary.
CBM-zero and FACE supply concept-level evidence through different mechanisms,
so \texttt{InterpretData} must state what the concepts mean and
\texttt{InterpretModel} or \texttt{PostHocInterpretation} must state how they
relate to predictions \citep{wang2025cbmzero,bhusal2025face}. FaithLM tests
whether a generated explanation tracks model behaviour, while FACT-E and
LIBERTy provide intervention-based evaluation of reasoning or concept claims
\citep{chuang2026faithlm,sun2026facte,toker2026liberty}. These tests are
additional evidence; fluent \texttt{Aggregation} alone does not establish
faithfulness. For a clinical audience, CHiRPE illustrates why domain mapping
and presentation should be adapted to recipients rather than inferred from
attribution scores alone \citep{fong2026chirpe}.

GFGE therefore provides a common procedural representation alongside the
classical axes. It makes method composition, stakeholder alignment, and
possible failure points explicit, while leaving fidelity and usefulness to
independent evaluation \citep{piculin2025userstudies}.

\section{Instantiating Existing Explanation Methods Using GFGE}
\label{sec:instantiation}

We now instantiate GFGE for established and recent method families. These
analytical mappings examine whether heterogeneous techniques can be described
with the same module roles; they are not empirical validation of explanation
quality. In each case, $\mathcal{P}$ denotes preprocessing for the predictive
model, while method-specific perturbation, search, or intervention is assigned
to an interpretation module.

\subsection{Post-Hoc Attribution Techniques: LIME and SHAP}

Post-hoc attribution methods such as LIME \citep{ribeiro2016should} and SHAP
\citep{lundberg2017unified} seek to explain the predictions of a model without
requiring access to its internal mechanisms. They address scenarios where the
interpretability of $\mathcal{M}$ is limited and explanation requires surrogate
or attribution-based tools.

\paragraph{LIME as an Instance of GFGE.}
The GFGE instantiation of LIME is as follows:
\begin{itemize}
    \item $\mathcal{M}$: any predictive model, including black-box neural models.
    \item $\mathcal{D}$: the instance to be explained.
    \item $\mathcal{P}$: preprocessing of the instance for $\mathcal{M}$.
    \item \texttt{InterpretData}: conversion of features into a human-readable form.
    \item \texttt{InterpretModel}: no internal evidence is required.
    \item \texttt{InterpretOutput}: the target prediction and queried outputs.
    \item \texttt{PostHocInterpretation}: generation of local perturbations and fitting
    of a sparse surrogate near $\mathcal{D}$.
    \item \texttt{Aggregation}: formatting of surrogate coefficients into a local
    explanation highlighting important features.
\end{itemize}

\paragraph{SHAP as an Instance of GFGE.}
SHAP integrates into GFGE in a similar fashion:
\begin{itemize}
    \item $\mathcal{D}$: the instance to be explained.
    \item $\mathcal{P}$: preprocessing of the instance for $\mathcal{M}$.
    \item \texttt{InterpretData}: record feature semantics and the background
    population needed to interpret the attribution.
    \item \texttt{InterpretModel}: no internal evidence is required for the
    model-agnostic implementation.
    \item \texttt{InterpretOutput}: the target prediction.
    \item \texttt{PostHocInterpretation}: evaluation of feature coalitions against an
    explicit background distribution, computation of the background-dependent
    reference value, and computation of Shapley values.
    \item \texttt{Aggregation}: representation of these contributions through ranked
    feature lists, bar plots, or natural language.
\end{itemize}

For a model-specific variant such as TreeSHAP \citep{lundberg2018consistent},
\texttt{InterpretModel} records the tree structure accessed by the procedure and
\texttt{PostHocInterpretation} computes contributions using that structure.

\subsection{Methods Built on Post-Hoc Attribution Models}

Beyond standalone post-hoc attribution methods, a growing body of research
constructs richer explanatory procedures by using attribution models as internal
signals. In these approaches, SHAP or similar techniques guide additional
reasoning steps such as contrastive reasoning, counterfactual generation, or the
aggregation of local explanations into global patterns. Examples include SHAP-guided counterfactual and contrastive explanations \citep{rathi2019generating},
Counterfactual Shapley Additive Explanations (CF-SHAP)
\citep{albini2022counterfactual}, methods that derive global feature importance by
aggregating local counterfactuals \citep{alfeo2023local}, and
SHAP-enhanced counterfactual explanations for recommendations
\citep{zhong2022shap}. These methods form a natural family within GFGE because
they combine post-hoc attribution with additional reasoning components.

\paragraph{SHAP-Enhanced Counterfactuals in GFGE.}
We illustrate this family using the recommender method from
\citep{zhong2022shap}.
\begin{itemize}
    \item $\mathcal{M}$: a recommendation model such as matrix factorization or a
    neural recommender.
    \item $\mathcal{D}$: the user item interaction to be explained.
    \item $\mathcal{P}$: preprocessing of the user--item interaction for the
    recommendation model.
    \item \texttt{InterpretData}: extraction of interpretable user and item features that
    might change in counterfactual scenarios.
    \item \texttt{InterpretModel}: optional inspection of available model structure;
    otherwise, no internal evidence is assumed.
    \item \texttt{InterpretOutput}: recording the model outputs obtained for
    evaluated counterfactual candidates.
    \item \texttt{PostHocInterpretation}: computation of Shapley values to identify the
    most influential features for the current recommendation, followed by a
    counterfactual search that prioritises modifications to these features.
    \item \texttt{Aggregation}: synthesis of SHAP attributions and the counterfactual
    into a unified explanation, for example  
    \emph{“Item A is recommended because features X and Y contributed strongly
    to its score. To obtain Item B instead, the following feature-level changes
    would be required.”}
\end{itemize}

This instantiation shows how GFGE captures hybrid explanation pipelines where
post-hoc attributions shape the search space of counterfactual reasoning.

\subsection{Model-Intrinsic Explanation Methods Based on Argumentation}

A different family of explainable approaches relies on intrinsically
interpretable models whose structure encodes argumentative relations. Examples
include A3R \citep{zhong2022a3r}, CA-FATA \citep{zhong2025argumentation}, and
argumentative explanations for interactive recommendations. In these methods,
explanations are produced directly by the model, not added post-hoc.

\paragraph{Argumentation-Based Recommendation Methods as Instances of GFGE.}
Their instantiation within GFGE is as follows:
\begin{itemize}
    \item $\mathcal{M}$: an argumentation-enhanced recommender model whose
    prediction is derived from evaluating competing arguments.
    \item $\mathcal{D}$: the user profile and candidate items.
    \item $\mathcal{P}$: transformation of user preferences and item attributes into
    argument premises.
    \item \texttt{InterpretData}: extraction of features that can function as premises or
    constraints in arguments.
    \item \texttt{InterpretModel}: retrieval of support and attack relations and the
    argument set used by the predictor.
    \item \texttt{InterpretOutput}: recording the recommendation and its score or
    decision; the active arguments remain model evidence in \texttt{InterpretModel}.
    \item \texttt{PostHocInterpretation}: not required, since explanations are embedded in
    the model itself.
    \item \texttt{Aggregation}: translation of argument structures into a human-readable
    explanation, often in natural language.
\end{itemize}

\subsection{Counterfactual and Recourse Methods}

Classical counterfactual methods search for inputs associated with an
alternative prediction, sometimes subject to diversity or feasibility
constraints \citep{wachter2018counterfactual,mothilal2020explaining,
poyiadzi2020face,ustun2019actionable}. In GFGE, \texttt{InterpretData} records which
features may change and which combinations are plausible; $q$ specifies the
desired outcome, while \texttt{InterpretOutput} records observed and candidate
model outputs; \texttt{PostHocInterpretation} searches for candidates;
and \texttt{Aggregation} reports the selected contrast with its assumptions.
This decomposition separates a prediction-changing input from a feasible
action in the world.

EECE provides a model-specific instantiation for random forests
\citep{zhang2025efficient}. \texttt{InterpretModel} exposes the decision-tree leaf
structure used to construct candidate regions, while
\texttt{PostHocInterpretation} searches those regions for counterfactuals.
\texttt{InterpretOutput} records the forest's responses to the evaluated
candidates. \texttt{InterpretData} states permitted feature changes and
plausibility constraints; \texttt{Aggregation} presents the selected candidate and
the properties actually assessed, such as validity, proximity, sparsity,
diversity, and actionability. The causal assumptions of backtracking
counterfactuals belong in the same data and evidence roles, rather than being
inferred from a changed prediction \citep{fatemi2025backtracking}.
Visual counterfactuals such as LeapFactual and V-CECE also populate these
roles, but require semantic interpretation of generated or edited images before
the alternative can be communicated meaningfully
\citep{cao2025leapfactual,spanos2025vcece}.

\subsection{Surrogate, Rule, Concept, and Prototype Methods}

A surrogate tree approximates another model: its rules arise from
post-hoc analysis of input--output behaviour
\citep{craven1996extracting,frosst2017distilling}. By contrast, a rule list
that performs the prediction itself exposes its active rule through model and
output interpretation; it needs no separate surrogate stage
\citep{wang2015falling,angelino2018learning}. The same visual artefact---a
rule---therefore occupies different GFGE roles depending on its source.

Concept and prototype methods require \texttt{InterpretData} to connect learned
features to domain concepts. For TCAV, \texttt{PostHocInterpretation} estimates a
concept's influence on a trained model; for ProtoPNet, \texttt{InterpretModel}
exposes prototype similarity used in prediction
\citep{kim2018interpretability,chen2019looks}. CBM-zero makes a pretrained
model's predictions accessible through an invertible concept mapping, while
FACE learns concept evidence with classifier supervision
\citep{wang2025cbmzero,bhusal2025face}. In the GFGE mapping of CBM-zero,
constructing the concept mapping belongs to PH because it is an additional
analytic construction over the pretrained predictor, rather than a
predictive mechanism originally used by the base model. When the converted
predictor is the object of inspection, reading its concept-to-output
mechanism belongs to IM. The evidence record must therefore identify which
predictor is being inspected; the PH construction and subsequent IM
inspection are distinct operations, as recorded in
Table~\ref{tab:same_cases}. \texttt{Aggregation} then selects concepts or
prototypes relevant to the recipient's question. Their semantic meaning
and predictive fidelity remain distinct evaluation claims.

\subsection{Language-Model and Recipient-Oriented Workflows}

A language model can render recorded evidence as a readable rationale through
\texttt{Aggregation} without thereby revealing its actual decision process.
When language generation is used within \texttt{Aggregation}, it is restricted
to rendering already recorded claims. Any newly introduced explanatory
proposition must first be recorded as a candidate claim and supported through
ID, IM, IO, or PH before it can be communicated as part of the explanation.
FaithLM uses
interventions to test and refine such accounts, adding evidence to
\texttt{PostHocInterpretation}; mechanistic circuit studies instead populate
\texttt{InterpretModel} with internal evidence
\citep{chuang2026faithlm,kim2025reasoningcircuits,arad2025circuits}.
CAuSE combines intervention-based analysis with natural-language explanation
of multimodal classifiers \citep{bandyopadhyay2026cause}. FACT-E and LIBERTy
are evaluation procedures for reasoning-chain and concept explanations;
they test claims generated by a method rather than serving as explanations by
themselves \citep{sun2026facte,toker2026liberty}.

For a clinician-facing workflow such as CHiRPE, \texttt{InterpretData} maps findings
to symptom domains, \texttt{PostHocInterpretation} contributes attribution evidence,
and \texttt{Aggregation} presents it in formats informed by clinicians
\citep{fong2026chirpe}. The intended audience is therefore a parameter of the
mapping, not merely a property of the final graphic or text. Recipient studies
must still test whether this presentation supports the intended task
\citep{piculin2025userstudies,mangold2026human}.

\subsection{Discussion}

The instantiations above show the sense in which GFGE unifies existing XAI
methods. Each method can be expressed as a configuration of the same
interpretative and communicative roles in
Equation~\ref{eq:gfge_mapping}, while retaining its own evidence source,
scope, and assumptions. This does not collapse attributions, counterfactuals,
rules, and generated rationales into equivalent claims. It makes their
different information flows and evidential requirements explicit.

Table~\ref{tab:gfge_instantiations} extends the worked examples to the major
families in the literature. ID, IM, IO, PH, and A denote
\texttt{InterpretData}, \texttt{InterpretModel},
\texttt{InterpretOutput}, \texttt{PostHocInterpretation}, and
\texttt{Aggregation}. The listed roles describe representative
implementations; particular methods can instantiate additional role nodes or repeat role types.

\begin{table}[H]
\centering\scriptsize
\caption{Representative XAI families within the common GFGE procedure.}
\label{tab:gfge_instantiations}
\begin{tabular}{@{}>{\raggedright\arraybackslash}p{2.65cm}>{\raggedright\arraybackslash}p{3.35cm}>{\raggedright\arraybackslash}p{4.15cm}@{}}
\toprule
\textbf{Family} & \textbf{Representative methods} & \textbf{GFGE roles and evidential issue} \\
\midrule
Feature attribution
& SHAP, TreeSHAP, Integrated Gradients, DeepLIFT, LRP, Grad-CAM
\citep{lundberg2017unified,lundberg2018consistent,sundararajan2017axiomatic,shrikumar2017learning,bach2015pixel,selvaraju2017grad}
& ID grounds features; IO records the prediction; PH computes contributions; A presents them. IM supplies structure for model-specific variants. \\
\midrule
Local and global surrogates
& LIME, surrogate trees, soft decision trees
\citep{ribeiro2016should,craven1996extracting,frosst2017distilling}
& IO gathers model responses; PH fits an approximation; A reports its rules or coefficients with their scope. \\
\midrule
Counterfactual and recourse
& Wachter CF, DiCE, FACE, EECE, BRACE
\citep{wachter2018counterfactual,mothilal2020explaining,poyiadzi2020face,zhang2025efficient,fatemi2025backtracking}
& ID records mutable, plausible changes; $q$ fixes the alternative; IO checks outputs; PH searches candidates; A communicates the contrast. EECE also uses IM for forest structure. \\
\midrule
Attribution-driven hybrid
& SHAP-guided CF, CF-SHAP, recommendation CF
\citep{rathi2019generating,albini2022counterfactual,zhong2022shap}
& PH composes attribution and search; ID constrains changes; IO records candidate model outputs; A combines findings and assumptions. \\
\midrule
Concept and prototype
& TCAV, ProtoPNet, ProtoTree, CBM-zero, FACE concept extraction
\citep{kim2018interpretability,chen2019looks,nauta2021neural,wang2025cbmzero,bhusal2025face}
& ID grounds concepts or prototypes; IM exposes intrinsic concept mechanisms; PH tests post-hoc influence; A presents selected evidence. \\
\midrule
Intrinsic rules and lists
& Falling Rule Lists, CORELS, decision sets
\citep{wang2015falling,angelino2018learning,lakkaraju2016interpretable}
& IM exposes predictive and active rules; IO records the result; ID grounds rule terms; A adapts the account to the question. \\
\midrule
Argumentation-based models
& A3R, CA-FATA, argumentative recommendation
\citep{zhong2022a3r,zhong2025argumentation,rago2018argumentation}
& ID grounds premises; IM exposes support, attack, and active arguments; IO records the result; A communicates the justification. \\
\midrule
Generated rationales
& FaithLM, CAuSE \citep{chuang2026faithlm,bandyopadhyay2026cause}
& IO records target behaviour and, when produced by $\mathcal{M}$, candidate rationales; PH tests explanatory claims; A renders supported, recorded claims. \\
\midrule
Mechanistic analyses
& Circuit localisation, reasoning circuits
\citep{arad2025circuits,kim2025reasoningcircuits}
& IM identifies internal components; PH may test causal relevance; A states supported mechanistic claims. \\
\midrule
Recipient-oriented workflows
& Contrastive accounts, CHiRPE
\citep{bucinca2025contrastive,fong2026chirpe}
& ID supplies domain meanings; $q$ selects a contrast, IO records corresponding outputs, and A adapts the account to the audience. \\
\bottomrule
\end{tabular}
\end{table}

The table exposes regularities that a family label does not. An attribution
and a surrogate both use PH, yet one reports contributions and the other an
approximation. A counterfactual adds an alternative outcome and constraints
on admissible changes; attribution-guided variants compose those operations.
A transparent rule or argumentation model may obtain reasons from IM without
PH, but still needs A to answer a recipient's question. An LLM rationale
can be readable without establishing that it reflects model computation.
GFGE's commonality is procedural; evidence and validity remain
method-dependent.

\paragraph{An Author-Coded Boundary Audit Against Nearby Frameworks.}
To make the claimed difference inspectable, Table~\ref{tab:same_cases}
applies the same three external methods to the taxonomy of
Schwalbe and Finzel~\citep{schwalbe2024taxonomy}, the roadmap of
Wang et al.~\citep{wang2024roadmap}, and
GFGE. The first identifies relevant method traits, including interactivity
and human actors; the second connects stakeholder needs, lifecycle stage,
questions, and methods. We compare the information each \emph{published
representation requires a user to specify}, not what could be added in an
extension. The GFGE coding applies Table~\ref{tab:module_contracts} to the
method descriptions; it is our analysis, not a result reported by those
papers. Each case tests a different boundary: learned concept structure
versus its post-hoc construction, calibration of perturbation evidence, and
generation versus intervention testing of a rationale.

\begin{table}[H]
\centering\scriptsize
\caption{The same methods viewed through two related frameworks and GFGE.
Entries describe what each representation foregrounds; ``unspecified'' means
the cited method does not supply that GFGE operation in the present analysis.}
\label{tab:same_cases}
\begin{tabular}{@{}>{\raggedright\arraybackslash}p{1.55cm}>{\raggedright\arraybackslash}p{2.35cm}>{\raggedright\arraybackslash}p{2.55cm}>{\raggedright\arraybackslash}p{4.35cm}@{}}
\toprule
\textbf{Method} & \textbf{Method-trait taxonomy} & \textbf{Stakeholder roadmap} & \textbf{GFGE trace and unresolved claim} \\
\midrule
CBM-zero \citep{wang2025cbmzero}
& Concept-based, model-access-dependent representation; local concept claims can be distinguished from global model traits.
& Can pair a concept-level ``what/why'' question with an audience and stage.
& PH constructs an invertible concept mapping; IM reads the converted predictor's concept-to-output mechanism; IO records prediction preservation; ID must ground concept meaning. Recipient-specific A is unspecified. The order PH$\to$IM and the semantic gap are explicit. \\
\midrule
ReCalX \citep{decker2025recalx}
& Perturbation-based, post-hoc analysis with reliability as a relevant method property.
& Can identify a debugging need and method for a prediction question.
& IO records scores under perturbation; PH checks calibration, adjusts the queried evidence, then recomputes an explanation. A downstream account and its audience are unspecified. An attribution without the calibration assumption remains a flagged claim. \\
\midrule
FaithLM \citep{chuang2026faithlm}
& Language explanation with iterative intervention-based assessment.
& Can connect a recipient's question to a language explanation method.
& IO records rationales produced by $\mathcal{M}$; PH intervenes to test faithfulness and inform revision; A renders supported, recorded claims. This is a method-internal loop, not observed recipient feedback. Recipient understanding remains unspecified. \\
\bottomrule
\end{tabular}
\end{table}

The difference is not that these frameworks cannot mention evidence or
stakeholders. It is that GFGE requires the source and operation of each
communicated claim, the order of transformations, and a mark for any missing
semantic or recipient-facing step. The three rows demonstrate that this
contract can reveal boundary choices and omissions in selected cases. They do
not show that GFGE is uniquely expressive or that another framework could
not be extended to hold the same information.

\paragraph{Reproducibility and Remaining Validation.}
The audit is specified as a coding exercise that can be independently repeated: fix $q$, $u$, and $c$; read
the method's published operations in order; assign each operation only when
its necessary condition in Table~\ref{tab:module_contracts} holds; record
source, assumptions, and output claim; mark unsupported or absent roles as
unresolved; and retain disagreements rather than resolve them by an
``other'' category. This paper reports one author-coded pass over three
purposively chosen external cases, with no annotator agreement statistic and
no user study. An independent test should freeze these rules, sample unfamiliar
methods by a declared selection rule, use at least two independent coders, and
report role-level agreement, unresolved cases, and revisions to the codebook.

\paragraph{Failure Cases Exposed by the IES--GFGE Mapping.}
First, a LIME or SHAP attribution to engineered or latent features may be
mathematically sound yet unintelligible to a user. The weak link is ID's
mapping from model variables to recipient concepts: analyst sense-reading
has occurred, but sense-giving is inadequately supported. ReCalX adds that
PH may itself rest on unreliable predictions for perturbed inputs
\citep{decker2025recalx}.

Second, an intrinsically interpretable linear model, rule list, or
argumentation model can expose its structure through IM without answering
a contrastive question. That requires $q$ to identify the alternative,
IO to record its model output, and A to communicate it. Likewise, a counterfactual that changes a model score
is not necessarily an actionable intervention: ID must record mutability
and causal constraints \citep{fatemi2025backtracking}.

Third, a fluent generated rationale may omit the information used by a
language model. PH can supply intervention evidence, as in FaithLM, while
FACT-E and LIBERTy test distinct faithfulness claims
\citep{chuang2026faithlm,sun2026facte,toker2026liberty}. A passed test does
not make all explanations of the same family faithful.

Table~\ref{tab:gfge_vs_taxonomy} shows how this analysis retains conventional
labels while making the additional operations and gaps visible.

\begin{table}[H]
\centering\scriptsize
\caption{Explanatory gaps left implicit by taxonomic labels.}
\label{tab:gfge_vs_taxonomy}
\begin{tabular}{@{}>{\raggedright\arraybackslash}p{2.65cm}>{\raggedright\arraybackslash}p{2.45cm}>{\raggedright\arraybackslash}p{4.95cm}@{}}
\toprule
\textbf{Workflow} & \textbf{Usual label} & \textbf{GFGE diagnosis} \\
\midrule
LIME/SHAP on latent features
& Local, post-hoc
& PH can compute scores while ID lacks a user-understandable mapping; A should state that limitation. \\
\midrule
Attribution-guided counterfactual
& Attribution or counterfactual
& PH composes two operations; ID constrains changes, $q$ specifies the target, IO checks it, and A joins the claims. \\
\midrule
EECE or causal recourse
& Counterfactual, post-hoc
& IM may supply forest structure; ID records feasible changes; PH searches candidates; A reports assumptions. \\
\midrule
Intrinsic rule or argumentation model
& Inherently interpretable
& IM exposes reasoning structure, but IO selects the relevant contrast and A addresses the recipient. \\
\midrule
Concept/prototype hybrid
& Concept or prototype based
& ID grounds semantics, IM or PH identifies influence, and A communicates the supported finding. \\
\midrule
Generated LLM rationale
& Natural-language explanation
& New propositions are recorded as candidates and require evidence in ID, IM, IO, or PH; A renders supported claims. PH can test faithfulness to model behaviour. \\
\bottomrule
\end{tabular}
\end{table}

These mappings are illustrative. They establish neither completeness nor
comparative superiority. A stronger evaluation would ask independent
annotators to map unfamiliar methods to GFGE and measure agreement and
coverage; test evidential claims within each module; and study whether the
resulting accounts improve understanding or decisions for intended
recipients. These three levels of evidence should remain separate.

\section{Conclusion}

This paper introduced GFGE to unify the description of explanation
generation across XAI methods. Grounded in interpretative frameworks and
the complementary activities of sense-reading and sense-giving, the
Interpret/Explain Schema connects an analyst's interpretation of system
evidence to the communication of an account and its interpretation by a
recipient. GFGE operationalises this relation through data interpretation,
model interpretation, output interpretation, optional post-hoc analysis,
and aggregation, with the explanatory question, audience, and context
made explicit.

The method instantiations show how intrinsic, post-hoc, and hybrid
approaches can be expressed through this shared vocabulary while retaining
their distinct evidence sources, assumptions, and scope. Role-typed
operation graphs capture method-specific dependencies and repeated roles;
claim-level records connect communicated explanations to the operations
and evidence that support them. This representation makes the contribution
of each operation explicit, including the boundary between constructing
evidence and selecting and rendering supported claims. It provides a
common basis for comparing explanation procedures, examining hybrid
workflows, and identifying unresolved explanatory dependencies relative
to a specified question and declared evidential requirements.

Future work should assess the reproducibility of the coding contract
through independent mappings of unfamiliar methods, examine the support
for explanatory claims through method-specific evidence checks, and
evaluate recipient understanding and use through user studies. GFGE's
central contribution is a procedural account that brings evidence,
interpretation, and communication into a common framework for XAI.
This account supports the analysis of how existing methods contribute
to an explanation for a particular question and recipient.

\bibliographystyle{splncs04}
\bibliography{references} 

@article{Samek2021Review,
  author    = {Samek, Wojciech and Montavon, Gr{\'e}goire and Lapuschkin, Sebastian and Anders, Christof J. and M{\"u}ller, Klaus-Robert},
  title     = {Explaining Deep Neural Networks and Beyond: A Review of Methods},
  journal   = {Proceedings of the IEEE},
  year      = {2021},
  volume    = {109},
  number    = {3},
  pages     = {247--278},
  doi       = {10.1109/JPROC.2021.3060483}
}

@article{zhong2025argumentation,
  title={Argumentation meets matrix factorization: A dual perspective for explainable recommendations},
  author={Zhong, Jinfeng and Negre, Elsa},
  journal={Applied Soft Computing},
  volume={183},
  pages={113652},
  year={2025},
  doi={10.1016/j.asoc.2025.113652},
  publisher={Elsevier}
}

@article{Vilone2021Notions,
  author    = {Vilone, Giulia and Longo, Luca},
  title     = {Notions of Explainability and Evaluation Approaches for Explainable Artificial Intelligence},
  journal   = {Information Fusion},
  year      = {2021},
  volume    = {76},
  pages     = {89--106},
  doi       = {10.1016/j.inffus.2021.05.009}
}

@inproceedings{Bhatt2020,
  author    = {Bhatt, Umang and Xiang, Alice and Sharma, Shubham and Weller, Adrian and Taly, Ankur and Jia, Yunhan and Ghosh, Joydeep and Puri, Ruchir and Moura, Jos{\'e} M. F. and Eckersley, Peter},
  title     = {Explainable Machine Learning in Deployment},
  booktitle = {Proceedings of the 2020 Conference on Fairness, Accountability, and Transparency},
  year      = {2020},
  pages     = {648--657},
  doi       = {10.1145/3351095.3375624}
}

@article{Wachter2017,
  author    = {Wachter, Sandra and Mittelstadt, Brent and Floridi, Luciano},
  title     = {Why a right to explanation of automated decision-making does not exist in the general data protection regulation},
  journal   = {International Data Privacy Law},
  year      = {2017},
  volume    = {7},
  number    = {2},
  pages     = {76--99},
  doi       = {10.1093/idpl/ipx005}
}

@misc{EUAIAct2024,
  author    = {{European Parliament and Council of the European Union}},
  title     = {Regulation ({EU}) 2024/1689 Laying Down Harmonised Rules on Artificial Intelligence},
  year      = {2024},
  note      = {{Artificial Intelligence Act}, Official Journal of the European Union, 12 July 2024},
  url       = {https://eur-lex.europa.eu/eli/reg/2024/1689/oj}
}

@inproceedings{fatemi2025backtracking,
  author    = {Fatemi, Pouria and Sharifian, Ehsan and Yassaee, Mohammad Hossein},
  title     = {A New Approach to Backtracking Counterfactual Explanations: A Unified Causal Framework for Efficient Model Interpretability},
  booktitle = {Proceedings of the 42nd International Conference on Machine Learning},
  series    = {Proceedings of Machine Learning Research},
  volume    = {267},
  pages     = {16273--16285},
  year      = {2025},
  url       = {https://proceedings.mlr.press/v267/fatemi25a.html}
}

@inproceedings{arad2025circuits,
  author    = {Arad, Dana and Belinkov, Yonatan and Chen, Hanjie and Kim, Najoung and Mohebbi, Hosein and Mueller, Aaron and Sarti, Gabriele and Tutek, Martin},
  title     = {Findings of the {BlackboxNLP} 2025 Shared Task: Localizing Circuits and Causal Variables in Language Models},
  booktitle = {Proceedings of the 8th BlackboxNLP Workshop: Analyzing and Interpreting Neural Networks for NLP},
  pages     = {543--552},
  year      = {2025},
  doi       = {10.18653/v1/2025.blackboxnlp-1.32}
}

@inproceedings{chuang2026faithlm,
  author    = {Chuang, Yu-Neng and Wang, Guanchu and Chang, Chia-Yuan and Tang, Ruixiang and Zhong, Shaochen and Yang, Fan and Wen, Andrew and Du, Mengnan and Cai, Xuanting and Braverman, Vladimir and Hu, Xia},
  title     = {{{FaithLM}}: Towards Faithful Explanations for Large Language Models},
  booktitle = {Proceedings of the 19th Conference of the European Chapter of the Association for Computational Linguistics (Volume 1: Long Papers)},
  pages     = {3802--3824},
  year      = {2026},
  doi       = {10.18653/v1/2026.eacl-long.177}
}

@article{mohseni2021multidisciplinary,
	title={A multidisciplinary survey and framework for design and evaluation of explainable {AI} systems},
	author={Mohseni, Sina and Zarei, Niloofar and Ragan, Eric D},
	journal={ACM Transactions on Interactive Intelligent Systems (TiiS)},
	volume={11},
	number={3-4},
	pages={1--45},
	year={2021},
	publisher={ACM New York, NY}
}

@inproceedings{shrikumar2017learning,
	title={Learning important features through propagating activation differences},
	author={Shrikumar, Avanti and Greenside, Peyton and Kundaje, Anshul},
	booktitle={International Conference on Machine Learning},
	pages={3145--3153},
	year={2017},
	organization={PMLR}
}

@incollection{lewis1986causal,
  title={Causal Explanation},
  author={Lewis, David},
  booktitle={Philosophical Papers, Volume II},
  pages={214--240},
  publisher={Oxford University Press},
  year={1986}
}

@article{lombrozo2006structure,
	title={The structure and function of explanations},
	author={Lombrozo, Tania},
	journal={Trends in Cognitive Sciences},
	volume={10},
	number={10},
	pages={464--470},
	year={2006},
	publisher={Elsevier}
}

@book{josephson1996abductive,
	title={Abductive inference: Computation, philosophy, technology},
	author={Josephson, John R and Josephson, Susan G},
	year={1996},
	publisher={Cambridge University Press}
}

@article{guidotti2018survey,
	title={A survey of methods for explaining black box models},
	author={Guidotti, Riccardo and Monreale, Anna and Ruggieri, Salvatore and Turini, Franco and Giannotti, Fosca and Pedreschi, Dino},
	journal={ACM Computing Surveys (CSUR)},
	volume={51},
	number={5},
	pages={1--42},
	year={2018},
	publisher={ACM New York, NY, USA}
}

@inproceedings{selvaraju2017grad,
  title={{Grad-CAM}: Visual explanations from deep networks via gradient-based localization},
  author={Selvaraju, Ramprasaath R and Cogswell, Michael and Das, Abhishek and Vedantam, Ramakrishna and Parikh, Devi and Batra, Dhruv},
  booktitle={Proceedings of the IEEE International Conference on Computer Vision},
  pages={618--626},
  year={2017}
}

@book{malle2006mind,
	title={How the mind explains behavior: Folk explanations, meaning, and social interaction},
	author={Malle, Bertram F},
	year={2006},
	publisher={MIT Press}
}

@article{arrieta2020explainable,
  title={Explainable Artificial Intelligence ({XAI}): Concepts, taxonomies, opportunities and challenges toward responsible {AI}},
  author={Arrieta, Alejandro Barredo and D{\'\i}az-Rodr{\'\i}guez, Natalia and Del Ser, Javier and Bennetot, Adrien and Tabik, Siham and Barbado, Alberto and Garc{\'\i}a, Salvador and Gil-L{\'o}pez, Sergio and Molina, Daniel and Benjamins, Richard and Chatila, Raja and Herrera, Francisco},
  journal={Information Fusion},
  volume={58},
  pages={82--115},
  year={2020},
  doi={10.1016/j.inffus.2019.12.012},
  publisher={Elsevier}
}

@article{carvalho2019machine,
	title={Machine learning interpretability: A survey on methods and metrics},
	author={Carvalho, Diogo V and Pereira, Eduardo M and Cardoso, Jaime S},
	journal={Electronics},
	volume={8},
	number={8},
	pages={832},
	year={2019},
	publisher={MDPI}
}

@article{montavon2018methods,
	title={Methods for interpreting and understanding deep neural networks},
	author={Montavon, Gr{\'e}goire and Samek, Wojciech and M{\"u}ller, Klaus-Robert},
	journal={Digital Signal Processing},
	volume={73},
	pages={1--15},
	year={2018},
	publisher={Elsevier}
}

@article{doshi2017towards,
  title={Towards a rigorous science of interpretable machine learning},
  author={Doshi-Velez, Finale and Kim, Been},
  journal={arXiv preprint arXiv:1702.08608},
  year={2017}
}

@article{lipton2018mythos,
  title={The mythos of model interpretability: In machine learning, the concept of interpretability is both important and slippery.},
  author={Lipton, Zachary C},
  journal={Queue},
  volume={16},
  number={3},
  pages={31--57},
  year={2018},
  publisher={ACM New York, NY, USA}
}

@book{hall2019introduction,
  title={An introduction to machine learning interpretability},
  author={Hall, Patrick and Gill, Navdeep},
  year={2019},
  publisher={O'Reilly Media, Incorporated}
}

@inproceedings{zhong2021ai,
  title={{AI}: To interpret or to explain?},
  author={Zhong, Jinfeng and Negre, Elsa},
  booktitle={Congr{\`e}s Inforsid (INFormatique des ORganisations et Syst{\`e}mes d'Information et de D{\'e}cision)},
  pages={149--164},
  year={2021},
  note={{Dijon, France}. {HAL}: hal-03529203}
}

@article{miller2019explanation,
  title={Explanation in artificial intelligence: Insights from the social sciences},
  author={Miller, Tim},
  journal={Artificial Intelligence},
  volume={267},
  pages={1--38},
  year={2019},
  publisher={Elsevier}
}

@article{jones2011mental,
  title={Mental models: an interdisciplinary synthesis of theory and methods},
  author={Jones, Natalie A and Ross, Helen and Lynam, Timothy and Perez, Pascal and Leitch, Anne},
  journal={Ecology and Society},
  volume={16},
  number={1},
  year={2011},
  publisher={JSTOR}
}

@inproceedings{tsuchiya1993improving,
  title={Improving knowledge creation ability through organizational learning},
  author={Tsuchiya, Shigehisa},
  booktitle={ISMICK'93 Proceedings, International Symposium on the Management of Industrial and Corporate Knowledge},
  pages={87--95},
  year={1993}
}

@article{lundberg2017unified,
  title={A unified approach to interpreting model predictions},
  author={Lundberg, Scott M and Lee, Su-In},
  journal={Advances in Neural Information Processing Systems},
  volume={30},
  year={2017}
}

@inproceedings{ribeiro2016should,
  title={{``Why Should I Trust You?'': Explaining the Predictions of Any Classifier}},
  author={Ribeiro, Marco Tulio and Singh, Sameer and Guestrin, Carlos},
  booktitle={Proceedings of the 22nd ACM SIGKDD International Conference on Knowledge Discovery and Data Mining},
  pages={1135--1144},
  year={2016}
}

@article{rudin2019stop,
  title={Stop explaining black box machine learning models for high stakes decisions and use interpretable models instead},
  author={Rudin, Cynthia},
  journal={Nature Machine Intelligence},
  volume={1},
  number={5},
  pages={206--215},
  year={2019},
  publisher={Nature Publishing Group}
}

@article{despres1999knowledge,
	title={Knowledge management (s)},
	author={Despres, Charles and Chauvel, Daniele},
	journal={Journal of Knowledge Management},
	volume={3},
	number={2},
	pages={110--123},
	year={1999},
	publisher={MCB UP Ltd}
}

@article{wachter2018counterfactual,
  title={Counterfactual explanations without opening the black box: Automated decisions and the {GDPR}},
  author={Wachter, Sandra and Mittelstadt, Brent and Russell, Chris},
  journal={Harvard Journal of Law and Technology},
  volume={31},
  number={2},
  pages={841--887},
  year={2018}
}

@article{du2019techniques,
	title={Techniques for interpretable machine learning},
	author={Du, Mengnan and Liu, Ninghao and Hu, Xia},
	journal={Communications of the ACM},
	volume={63},
	number={1},
	pages={68--77},
	year={2019},
	publisher={ACM New York, NY, USA}
}

@article{gilpin2018explaining,
	title={Explaining explanations: An approach to evaluating interpretability of machine learning},
	author={Gilpin, Leilani H and Bau, David and Yuan, Ben Z and Bajwa, Ayesha and Specter, Michael and Kagal, Lalana},
	journal={arXiv preprint arXiv:1806.00069},
	year={2018},
	publisher={CoRR}
}

@article{hoffman2018metrics,
	title={Metrics for explainable {AI}: Challenges and prospects},
	author={Hoffman, Robert R and Mueller, Shane T and Klein, Gary and Litman, Jordan},
	journal={arXiv preprint arXiv:1812.04608},
	year={2018}
}

@inproceedings{mittelstadt2019explaining,
title={Explaining explanations in {AI}},
author={Mittelstadt, Brent and Russell, Chris and Wachter, Sandra},
booktitle={Proceedings of the Conference on Fairness, Accountability, and Transparency},
pages={279--288},
year={2019}
}

@book{molnar2020interpretable,
  title={Interpretable machine learning},
  author={Molnar, Christoph},
  year={2020},
  publisher={Lulu. com}
}

@inproceedings{ustun2019actionable,
  title={Actionable recourse in linear classification},
  author={Ustun, Berk and Spangher, Alexander and Liu, Yang},
  booktitle={Proceedings of the Conference on Fairness, Accountability, and Transparency},
  pages={10--19},
  year={2019},
  doi={10.1145/3287560.3287566}
}

@article{frosst2017distilling,
  title={Distilling a neural network into a soft decision tree},
  author={Frosst, Nicholas and Hinton, Geoffrey},
  journal={arXiv preprint arXiv:1711.09784},
  year={2017}
}

@inproceedings{poyiadzi2020face,
  title={{FACE: Feasible and Actionable Counterfactual Explanations}},
  author={Poyiadzi, Rafael and Sokol, Kacper and Santos-Rodriguez, Raul and De Bie, Tijl and Flach, Peter},
  booktitle={Proceedings of the AAAI/ACM Conference on AI, Ethics, and Society},
  pages={344--350},
  year={2020},
  doi={10.1145/3375627.3375850}
}

@inproceedings{mothilal2020explaining,
  title={Explaining machine learning classifiers through diverse counterfactual explanations},
  author={Mothilal, Ramaravind K and Sharma, Amit and Tan, Chenhao},
  booktitle={Proceedings of the 2020 Conference on Fairness, Accountability, and Transparency},
  pages={607--617},
  year={2020}
}

@article{angelino2018learning,
  title={Learning certifiably optimal rule lists for categorical data},
  author={Angelino, Elaine and Larus-Stone, Nicholas and Alabi, Daniel and Seltzer, Margo and Rudin, Cynthia},
  journal={Journal of Machine Learning Research},
  volume={18},
  number={234},
  pages={1--78},
  year={2018}
}

@inproceedings{wang2015falling,
  title={Falling rule lists},
  author={Wang, Fulton and Rudin, Cynthia},
  booktitle={Artificial Intelligence and Statistics},
  pages={1013--1022},
  year={2015},
  organization={PMLR}
}

@inproceedings{lakkaraju2016interpretable,
  title={Interpretable decision sets: A joint framework for description and prediction},
  author={Lakkaraju, Himabindu and Bach, Stephen H and Leskovec, Jure},
  booktitle={Proceedings of the 22nd ACM SIGKDD International Conference on Knowledge Discovery and Data Mining},
  pages={1675--1684},
  year={2016}
}

@book{craven1996extracting,
  title={Extracting comprehensible models from trained neural networks},
  author={Craven, Mark William},
  year={1996},
  publisher={The University of Wisconsin-Madison}
}

@article{bach2015pixel,
  title={On pixel-wise explanations for non-linear classifier decisions by layer-wise relevance propagation},
  author={Bach, Sebastian and Binder, Alexander and Montavon, Gr{\'e}goire and Klauschen, Frederick and M{\"u}ller, Klaus-Robert and Samek, Wojciech},
  journal={{PLOS ONE}},
  volume={10},
  number={7},
  pages={e0130140},
  year={2015},
  publisher={Public Library of Science San Francisco, CA USA}
}

@article{alfeo2023local,
  title={From local counterfactuals to global feature importance: efficient, robust, and model-agnostic explanations for brain connectivity networks},
  author={Alfeo, Antonio Luca and Zippo, Antonio G and Catrambone, Vincenzo and Cimino, Mario GCA and Toschi, Nicola and Valenza, Gaetano},
  journal={Computer Methods and Programs in Biomedicine},
  volume={236},
  pages={107550},
  year={2023},
  publisher={Elsevier}
}

@inproceedings{albini2022counterfactual,
  title={Counterfactual shapley additive explanations},
  author={Albini, Emanuele and Long, Jason and Dervovic, Danial and Magazzeni, Daniele},
  booktitle={Proceedings of the 2022 ACM Conference on Fairness, Accountability, and Transparency},
  pages={1054--1070},
  year={2022},
  doi={10.1145/3531146.3533168}
}

@inproceedings{kim2018interpretability,
  title={Interpretability beyond feature attribution: Quantitative testing with concept activation vectors ({TCAV})},
  author={Kim, Been and Wattenberg, Martin and Gilmer, Justin and Cai, Carrie and Wexler, James and Viegas, Fernanda and Sayres, Rory},
  booktitle={International Conference on Machine Learning},
  pages={2668--2677},
  year={2018},
  organization={PMLR}
}

@inproceedings{nauta2021neural,
  title={Neural prototype trees for interpretable fine-grained image recognition},
  author={Nauta, Meike and Van Bree, Ron and Seifert, Christin},
  booktitle={Proceedings of the IEEE/CVF Conference on Computer Vision and Pattern Recognition},
  pages={14933--14943},
  year={2021}
}

@article{chen2019looks,
  title={This looks like that: deep learning for interpretable image recognition},
  author={Chen, Chaofan and Li, Oscar and Tao, Daniel and Barnett, Alina and Rudin, Cynthia and Su, Jonathan K},
  journal={Advances in Neural Information Processing Systems},
  volume={32},
  year={2019}
}

@article{rathi2019generating,
  title={Generating counterfactual and contrastive explanations using {SHAP}},
  author={Rathi, Shubham},
  journal={arXiv preprint arXiv:1906.09293},
  year={2019}
}

@inproceedings{zhong2022shap,
  title={{SHAP}-enhanced counterfactual explanations for recommendations},
  author={Zhong, Jinfeng and Negre, Elsa},
  booktitle={Proceedings of the 37th ACM/SIGAPP Symposium on Applied Computing},
  pages={1365--1372},
  year={2022},
  doi={10.1145/3477314.3507029}
}

@inproceedings{zhong2022a3r,
	title={{A$^3$R}: Argumentative explanations for recommendations},
	author={Zhong, Jinfeng and Negre, Elsa},
	booktitle={2022 IEEE 9th International Conference on Data Science and Advanced Analytics (DSAA)},
	pages={1--9},
	year={2022},
	doi={10.1109/DSAA54385.2022.10032419},
	organization={IEEE}
}

@book{arduin2015information,
  title={Information and knowledge system},
  author={Arduin, Pierre-Emmanuel and Grundstein, Michel and Rosenthal-Sabroux, Camille},
  year={2015},
  publisher={John Wiley \& Sons}
}

@inproceedings{sundararajan2017axiomatic,
  title={Axiomatic attribution for deep networks},
  author={Sundararajan, Mukund and Taly, Ankur and Yan, Qiqi},
  booktitle={International Conference on Machine Learning},
  pages={3319--3328},
  year={2017},
  organization={PMLR}
}

@article{zhang2025efficient,
  title={Efficient and Effective Counterfactual Explanations for Random Forests},
  author={Zhang, Haifei and Zhong, Jinfeng},
  journal={Expert Systems with Applications},
  volume={293},
  pages={128661},
  year={2025},
  doi={10.1016/j.eswa.2025.128661},
  publisher={Elsevier}
}

@article{wang2024roadmap,
  author={Wang, Ziming and Huang, Changwu and Yao, Xin},
  title={A Roadmap of Explainable Artificial Intelligence: Explain to Whom, When, What and How?},
  journal={ACM Transactions on Autonomous and Adaptive Systems},
  volume={19},
  number={4},
  pages={20:1--20:40},
  year={2024},
  doi={10.1145/3702004}
}

@article{elsayed2026taxonomy,
  author={Elsayed, Abdelrahman and Wu, Yuandi and Yawney, John and Gadsden, Stephen Andrew},
  title={Explainable artificial intelligence: An expanded taxonomy by mechanism and interpretability perspectives},
  journal={Intelligent Systems with Applications},
  pages={200709},
  year={2026},
  doi={10.1016/j.iswa.2026.200709}
}

@inproceedings{cai2026actionable,
  author={Cai, Jie and Rahman, Mafizur and Enouen, James and Qian, Lijun and Liu, Yan},
  title={A Survey on Actionable Interpretability in Large Language Models},
  booktitle={Proceedings of the Thirty-Fifth International Joint Conference on Artificial Intelligence},
  pages={7787--7797},
  year={2026},
  doi={10.24963/ijcai.2026/865},
  url={https://www.ijcai.org/proceedings/2026/865}
}

@inproceedings{rago2018argumentation,
  title={Argumentation-based recommendations: Fantastic explanations and how to find them},
  author={Rago, Antonio and Cocarascu, Oana and Toni, Francesca},
  booktitle={Proceedings of IJCAI},
  pages={1949--1955},
  year={2018}
}

@article{lundberg2018consistent,
  title={Consistent individualized feature attribution for tree ensembles},
  author={Lundberg, Scott M and Erion, Gabriel G and Lee, Su-In},
  journal={arXiv preprint arXiv:1802.03888},
  year={2018}
}

@inproceedings{decker2025recalx,
  author    = {Decker, Thomas and Tresp, Volker and Buettner, Florian},
  title     = {Improving Perturbation-based Explanations by Understanding the Role of Uncertainty Calibration},
  booktitle = {Advances in Neural Information Processing Systems 38},
  year      = {2025},
  doi       = {10.52202/085713-0226},
}

@inproceedings{cao2025leapfactual,
  author    = {Cao, Zhuo and Zhao, Xuan and Krieger, Lena and Scharr, Hanno and Assent, Ira},
  title     = {{LeapFactual}: Reliable Visual Counterfactual Explanation Using Conditional Flow Matching},
  booktitle = {Advances in Neural Information Processing Systems 38},
  year      = {2025},
  doi       = {10.52202/085713-3961},
}

@inproceedings{spanos2025vcece,
  author    = {Spanos, Nikolaos and Lymperaiou, Maria and Filandrianos, Giorgos and Thomas, Konstantinos and Voulodimos, Athanasios and Stamou, Giorgos},
  title     = {{V-CECE}: Visual Counterfactual Explanations via Conceptual Edits},
  booktitle = {Advances in Neural Information Processing Systems 38},
  year      = {2025},
  doi       = {10.52202/085713-2493},
}

@inproceedings{wang2025cbmzero,
  author    = {Wang, Zhenzhen and Popel, Aleksander and Sulam, Jeremias},
  title     = {Concept Bottleneck Model with Zero Performance Loss},
  booktitle = {Conference on Parsimony and Learning},
  series    = {Proceedings of Machine Learning Research},
  volume    = {280},
  pages     = {433--461},
  year      = {2025},
  url       = {https://proceedings.mlr.press/v280/wang25b.html}
}

@inproceedings{bhusal2025face,
  author    = {Bhusal, Dipkamal and Clifford, Michael and Rampazzi, Sara and Rastogi, Nidhi},
  title     = {{FACE}: Faithful Automatic Concept Extraction},
  booktitle = {Advances in Neural Information Processing Systems 38},
  year      = {2025},
  doi       = {10.52202/085713-1462},
}

@inproceedings{piculin2025userstudies,
  author    = {Pi{\v{c}}ulin, Matej and Petek, Bernarda and Ograjen{\v{s}}ek, Irena and Strumbelj, Erik},
  title     = {Position: Explainable {{AI}} Cannot Advance Without Better User Studies},
  booktitle = {Proceedings of the 42nd International Conference on Machine Learning},
  series    = {Proceedings of Machine Learning Research},
  volume    = {267},
  pages     = {81977--81994},
  year      = {2025},
  url       = {https://proceedings.mlr.press/v267/piculin25a.html}
}

@inproceedings{bucinca2025contrastive,
  author    = {Bu{\c{c}}inca, Zana and Swaroop, Siddharth and Paluch, Amanda E. and Doshi-Velez, Finale and Gajos, Krzysztof Z.},
  title     = {Contrastive Explanations That Anticipate Human Misconceptions Can Improve Human Decision-Making Skills},
  booktitle = {Proceedings of the 2025 CHI Conference on Human Factors in Computing Systems},
  articleno = {1024},
  pages     = {1--25},
  year      = {2025},
  doi       = {10.1145/3706598.3713229}
}

@inproceedings{fong2026chirpe,
  author    = {Fong, Stephanie and Wang, Zimu and Oliveira, Guilherme C. and Zhao, Xiangyu and Jiang, Yiwen and Liu, Jiahe and Colton, Beau-Luke and Woods, Scott W. and Shenton, Martha and Nelson, Barnaby and Ge, Zongyuan and Dwyer, Dominic},
  title     = {{{CHiRPE}}: A Step Towards Real-World Clinical {NLP} with Clinician-Oriented Model Explanations},
  booktitle = {Proceedings of the 19th Conference of the European Chapter of the Association for Computational Linguistics (Volume 2: Short Papers)},
  pages     = {646--658},
  year      = {2026},
  doi       = {10.18653/v1/2026.eacl-short.46}
}

@inproceedings{kim2025reasoningcircuits,
  author    = {Kim, Geonhee and Valentino, Marco and Freitas, Andre},
  title     = {Reasoning Circuits in Language Models: A Mechanistic Interpretation of Syllogistic Inference},
  booktitle = {Findings of the Association for Computational Linguistics: ACL 2025},
  pages     = {10074--10095},
  year      = {2025},
  doi       = {10.18653/v1/2025.findings-acl.525}
}

@article{bandyopadhyay2026cause,
  author    = {Bandyopadhyay, Dibyanayan and Bhattacharjee, Soham and Hasanuzzaman, Mohammed and Ekbal, Asif},
  title     = {{{CAuSE}}: Decoding Multimodal Classifiers using Faithful Natural Language Explanation},
  journal   = {Transactions of the Association for Computational Linguistics},
  volume    = {14},
  pages     = {829--851},
  year      = {2026},
  doi       = {10.1162/tacl.a.686}
}

@inproceedings{sun2026facte,
  author    = {Sun, Yuxi and Zuo, Aoqi and Xie, Haotian and Gao, Wei and Gong, Mingming and Ma, Jing},
  title     = {{{FACT-E}}: Causality-Inspired Evaluation for Trustworthy Chain-of-Thought Reasoning},
  booktitle = {Findings of the Association for Computational Linguistics: ACL 2026},
  pages     = {20275--20293},
  year      = {2026},
  doi       = {10.18653/v1/2026.findings-acl.1014}
}

@inproceedings{toker2026liberty,
  author    = {Toker, Gilat and Calderon, Nitay and Amosy, Ohad and Reichart, Roi},
  title     = {{{LIBERTy}}: A Causal Framework for Benchmarking Concept-Based Explanations of {LLM}s with Structural Counterfactuals},
  booktitle = {Findings of the Association for Computational Linguistics: ACL 2026},
  pages     = {8149--8186},
  year      = {2026},
  doi       = {10.18653/v1/2026.findings-acl.399}
}

@article{schwalbe2024taxonomy,
  author  = {Schwalbe, Gesina and Finzel, Bettina},
  title   = {A Comprehensive Taxonomy for Explainable Artificial Intelligence: A Systematic Survey of Surveys on Methods and Concepts},
  journal = {Data Mining and Knowledge Discovery},
  volume  = {38},
  pages   = {3043--3101},
  year    = {2024},
  doi     = {10.1007/s10618-022-00867-8}
}

@article{cousineau2025taxonomy,
  author  = {Cousineau, Carter and Dara, Rozita},
  title   = {Interpretability Taxonomy: An Approach for Artificial Intelligence Developers Technique Selection},
  journal = {Discover Data},
  volume  = {3},
  pages   = {60},
  year    = {2025},
  doi     = {10.1007/s44248-025-00096-6}
}

@article{mangold2026human,
  author  = {Mangold, Aline and Zietz, Juliane and Weinhold, Susanne and Pannasch, Sebastian},
  title   = {On the Design and Evaluation of Human-Centered Explainable {AI} Systems: A Systematic Review and Taxonomy},
  journal = {Advances in Human-Computer Interaction},
  volume  = {2026},
  pages   = {8450102},
  year    = {2026},
  doi     = {10.1155/ahci/8450102}
}

\end{document}